\documentclass{article}

\usepackage{microtype}
\usepackage{graphicx}
\usepackage{subcaption}
\usepackage{booktabs}
\usepackage{hyperref}

\usepackage{algorithm}
\usepackage{algpseudocode}

\usepackage[preprint]{icml2026}
\usepackage{amsmath}
\usepackage{amssymb}
\usepackage{mathtools}
\usepackage{amsthm}
\usepackage[capitalize,noabbrev]{cleveref}
\usepackage{enumitem}
\usepackage{booktabs}
\usepackage{multirow}
\usepackage{svg}

\theoremstyle{plain}
\newtheorem{theorem}{Theorem}[section]

\newtheorem{corollary}{Corollary}[section]
\theoremstyle{definition}
\newtheorem{definition}{Definition}[section]
\newtheorem{assumption}{Assumption}[section]
\theoremstyle{remark}

\usepackage[textsize=tiny]{todonotes}

\begin{document}

\twocolumn[
  \icmltitle{When Sharpening Becomes Collapse: Sampling Bias and Semantic Coupling in RL with Verifiable Rewards}

  \icmlsetsymbol{equal}{*}
  \begin{icmlauthorlist}
    \icmlauthor{Mingyuan Fan}{ecnu}
    \icmlauthor{Weiguang Han}{ant}
    \icmlauthor{Daixin Wang}{ant}
    \icmlauthor{Cen Chen}{ecnu}
    \icmlauthor{Zhiqiang Zhang}{ant}
    \icmlauthor{Jun Zhou}{ant}
  \end{icmlauthorlist}

  \icmlaffiliation{ecnu}{School of Data Science\&Engineering, East China Normal University}
  \icmlaffiliation{ant}{Ant Group}

  \icmlcorrespondingauthor{Cen Chen}{cenchen@dase.ecnu.edu.cn}

  \icmlkeywords{Large Language Models, Reinforcement Learning}

  \vskip 0.3in
]



\printAffiliationsAndNotice{}  

\begin{abstract}
    Reinforcement Learning with Verifiable Rewards (RLVR) is a central paradigm for turning large language models (LLMs) into reliable problem solvers, especially in logic‑heavy domains.
    Despite its empirical success, it remains unclear whether RLVR elicits novel capabilities or merely sharpens the distribution over existing knowledge.
    We study this by formalizing over-sharpening, a phenomenon where the policy collapses onto limited modes, suppressing valid alternatives.
    At a high level, we discover finite-batch updates intrinsically bias learning toward sampled modes, triggering a collapse that propagates globally via semantic coupling.
    To mitigate this, we propose inverse-success advantage calibration to prioritize difficult queries and distribution-level calibration to diversify sampling via a memory network.
    Extensive evaluations show that our proposed methods can effectively maintain a healthy exploration-exploitation balance and achieve sustained accuracy gains compared to state-of-the-art methods.
\end{abstract}



\section{Introduction}
\label{sec1}

Reinforcement learning with verifiable rewards (RLVR) has become a cornerstone approach for specializing large language models (LLMs) to domains where correctness is checkable~\cite{deepseek_r1,qwen3}.
However, there is still no clear consensus on what RLVR fundamentally does to LLMs.
One line of work credits RLVR with unlocking new capabilities in LLMs, pointing to emergent chain‑of‑thought reasoning and strong performance on challenging benchmarks.
On the other hand, recent empirical studies~\cite{reasoning_inference,passk_better_base_model,passk_better_base_model2} paint a more conservative picture: when given sufficiently large sampling budgets, pre-trained models can often match or even surpass the performance of RL-fine-tuned counterparts.
These observations suggest that, rather than endowing models with new skills, RLVR may primarily act to sharpen the output distribution, making the model more confident in its existing knowledge.

We argue that these perspectives are not mutually exclusive.
From a distributional perspective, pre‑training and supervised fine‑tuning (SFT) provide the model with a rich, low‑dimensional manifold of knowledge and latent reasoning procedures.
RLVR then performs a search over this manifold, guided by verifiable rewards.
We formalize the core mechanism of this process as sharpening and distinguish between two critical regimes: \textit{\textbf{moderate and over-sharpening}}.
Moderate sharpening is beneficial, since it shifts probability mass toward high‑reward semantic modes, making the model more reliable under finite sampling.
However, if sharpening is too aggressive, the policy will collapse onto a narrow set of modes, prematurely discarding valid alternative strategies.
This restricts the model’s exploration to a small subset of the solution space, eventually stalling the discovery of novel reasoning chains and limiting overall capability growth.

This paper is most closely related to research on entropy collapse, a pervasive phenomenon in RLVR where the policy’s predictive distribution becomes prematurely concentrated.
In many respects, over-sharpening serves as a more rigorous and formal characterization of this collapse.
Typical examples to mitigate entropy collapse include token-level heuristics like selective updates of high-entropy tokens~\cite{rule_2080} or gradient truncation for highly confident ones~\cite{cov_update}, and batch-level interventions that discard degenerate samples~\cite{sample_filtering} or utilize curated synthetic data~\cite{sample_filtering2}.
Additionally, explicit entropy regularization and its variants (e.g., DAPO-style clipping~\cite{DAPO,qae} or direct objective penalties~\cite{entropy_reward,minimize_entropy}) are employed.
While effective in practice, these methods primarily treat entropy as a finite resource to be preserved through technical mitigations, leaving the underlying mechanisms of over-sharpening largely unexplored.
For example, it remains unclear whether over-sharpening is an inevitable byproduct of advancing model performance or a structural artifact of algorithmic bias.

We identify two mechanisms that drive over‑sharpening in RLVR, namely sampling bias and semantic coupling.
First, practical constraints force us to sample only a small number of responses per query during training.
These samples are typically drawn from high-probability modes, and therefore cover only a subset of all valid solutions.
This biases the model updates toward reinforcing those modes and gradually drives the policy to collapse on them.
Second, because model parameters are shared across inputs, updates from one sample affect the behavior on other samples, i.e., semantic coupling.
Thus, semantic coupling causes collapse at one point to propagate to semantically related queries.
Over time, the effective solution space for many inputs shrinks, the model gains little new information from additional RLVR steps, and collapse is further reinforced.
\cref{sec2} analyzes these mechanisms in detail, both theoretically and empirically.

To counteract these, we develop two complementary techniques that act on different parts of the RLVR pipeline.
At the learning level, we adjust how much each sampled trajectory influences the update so that easy, already‑solved queries do not dominate the optimization signal.
At the sampling level, we tune which trajectories are even considered for training so that the model is not repeatedly trained on the same high‑probability patterns and is encouraged to explore alternative reasoning paths.
Together, these mechanisms can effectively keep RLVR in a moderate sharpening regime, making the model more decisive where it matters, while avoiding the collapse of diverse, potentially useful behaviors.
Our contributions can be summarized as follows:
\begin{itemize}[leftmargin=*,topsep=1pt]
    \item \textbf{A formal framework of sharpening in RLVR.}
    We formalize the distinction between moderate and over‑sharpening and show that, in an ideal infinite‑sample setting, the standard RLVR objective does not suppress correct behaviors.
    We then analyze the realistic finite‑batch regime and reveal how sampling bias and shared parameters conspire to suppress unsampled modes and propagate collapse across semantically related queries.
    This yields a unified picture of when RLVR behaves as a benign sharpening mechanism and when it risks destroying useful diversity.
   
    \item \textbf{Two simple, general‑purpose calibration strategies for RLVR.}
    Building on this analysis, we introduce inverse‑success advantage calibration and distribution‑level calibration.
    The former one automatically down‑weights updates from easy, high‑success queries and emphasizes harder ones.
    The later one employs a memory network to reshape the rollout distribution to avoid over‑training on already dominant modes and to promote exploration of alternative reasoning patterns.
    Both techniques can be plugged into existing RLVR pipelines against over‑sharpening.

    \item \textbf{Extensive experiments over multiple LLMs and benchmark datasets.}
    The performance of the proposed methods over six mathematical reasoning benchmarks consistently outperforms state-of-the-art baselines across both AVG@8 and PASS@8 by about 2\%$\sim$3\%.
    We also show that our methods are compatible with existing entropy-regularization techniques (e.g., DAPO), delivering additive performance improvements.
\end{itemize}

\begin{table*}[t]
\centering
\caption{Summary of our theoretical findings.}
\label{tab:theory_summary}
\small
\begin{tabular}{@{}p{0.12\linewidth}p{0.88\linewidth}@{}}
\toprule
\textbf{Aspect} & \textbf{Key findings} \\ \midrule
\textbf{Infinite‑sample\newline Regime} &
The ideal KL‑regularized RLVR objective is inherently benign (moderate sharpening), where the probability of every correct mode is strictly increased (\cref{cor:moderate_exact}). \\ \midrule
\textbf{Sampling Bias\newline (Finite‑batch)} &
Collapse is triggered when the model becomes confident (i.e., $\Delta_\pi > 0$), causing the partition function to explode ($Z'(q) > 1$). This forces the suppression of all unsampled correct modes (\cref{thm:Z_bound}).  \\ \midrule
\textbf{Semantic\newline Coupling} &
 Updates do not remain local. The probability mass at $q'$ is disproportionately drawn toward the batch‑seen mode at $q$, thereby suppressing unseen but correct modes for $q$.  \\ \bottomrule
\end{tabular}
\end{table*}

\section{Related Work}
\label{sec5}

Since the release of OpenAI o1~\cite{open_o1} and DeepSeek-R1~\cite{deepseek_r1}, RLVR (or GRPO and its variants) has gained particular traction for enhancing the reasoning ability of LLMs.
A key driver of its effectiveness over SFT could be the explicit incorporation of negative samples~\cite{sample_filtering}.
To stabilize the often volatile training dynamics, a significant body of work focused on refining advantage estimation.
Prominent techniques include RLOO (Leave-One-Out)~\cite{RLOO}, DrGRPO~\cite{DrGRPO}, which modifies the variance term to improve stability, and Reinforce++ with baseline~\cite{reinforce++} that shifts from sample-level to global batch-level normalization.
Moreover, \citet{cov_update} identified an empirical law showing that performance gains are often traded for entropy, leading to predictable saturation.
To prevent such premature convergence, various entropy-management strategies have been proposed.
These include token-level heuristics targeting high-entropy forking tokens~\cite{rule_2080}, multi-temperature sampling for explicit exploration~\cite{multi_temp}, and entropy-aware clipping mechanisms~\cite{DAPO, qae}.
RENT~\cite{minimize_entropy} and RPN~\cite{uncertain_reward} demonstrate that minimizing entropy on confident chains or adding explicit regularization can also drive gains.
Despite these advancements, recent studies~\cite{reasoning_inference,passk_better_base_model,passk_better_base_model2} indicate that RLVR training can sometimes lead to a degradation in general capabilities.
Critically, the underlying mechanics of over-sharpening remain largely unexplored and this paper aims to bridge this gap.

\section{Analyzing Distribution Collapse} 
\label{sec2}

we first present that infinite‑sample RLVR always yields moderate sharpening, then show how finite‑batch sampling fundamentally alters this behavior and leads to over‑sharpening on other queries via semantic coupling.

\subsection{Problem Setup}
\label{sec2_1}

We consider a policy model $\pi_{\theta}$ initialized from a reference model $\pi_{\text{ref}}$.
For a query $q$, we coarsen the sequence‑level output space into $K_1$ mutually exclusive correct modes $\mathcal{O}^+ = \{o_1, \dots, o_{K_1}\}$ and $K_2$ incorrect modes $\mathcal{O}^-$.
Each mode can be viewed as an equivalence class of outputs that implement the same high‑level reasoning pattern or solution.
We focus on the RLVR setting with binary verifiable rewards, where correct modes share a non‑negative advantage $A_+ \ge 0$, and incorrect modes share $A_- < 0$.
\cref{def:sharpening} formalizes two possible behaviors of an updated policy relative to $\pi_{\text{ref}}$.
\begin{definition}
\label{def:sharpening}
A policy $\pi_{\text{new}}$ exhibits moderate sharpening if the probability of every correct mode increases (i.e., $\pi_{\text{new}}(o|q) \geq \pi_{\text{ref}}(o|q)$ for all $o \in \mathcal{O}^+$), and over-sharpening if the probability of any correct mode decreases.
\end{definition}
Moderate sharpening reflects a holistic improvement where the model gains confidence without forgetting valid reasoning patterns.
Over‑sharpening instead suppresses some correct modes in favor of others, which could be problematic because the suppressed patterns might be essential for solving other queries.
Moreover, as suppression intensifies, the policy degenerates into a deterministic distribution.
Thus, entropy collapse can be viewed as the asymptotic limit of severe over-sharpening.

\subsection{Sampling Bias}
\label{sec2_2}

\textbf{Infinite-sample sharpening behavior.}
We begin with the standard KL‑regularized RL objective:
\begin{equation}
\label{eq:full_obj}
\begin{split}
    \max_\theta \ \mathbb{E}_{q \sim \mathcal{D}} [ &\mathbb{E}_{o \sim \pi_{\theta}(\cdot|q)} [A(q, o)] \\
    &- \beta D_{KL}(\pi_\theta(\cdot|q) \| \pi_{ref}(\cdot|q)) ,
\end{split}
\end{equation}
where $\mathcal{D}$ is the query distribution, $A(q,o)$ is the advantage of $o$, and $\beta \geq 0$ is the regularization coefficient.
\cref{thm:optimal_exact} provides the closed‑form optimal solution of \cref{eq:full_obj}.

\begin{theorem}
\label{thm:optimal_exact}
The maximizer of \cref{eq:full_obj}, denoted as $\pi^*$, is given by
\begin{equation}
\label{eq:optimal_exact}
\pi^*(o|q) = \frac{1}{Z(q)}\, \pi_{\text{ref}}(o|q)\,
\exp\!\left( \frac{A(q,o)}{\beta} \right),
\end{equation}
where $Z(q)$ is the partition function ensuring normalization.
See \cref{appendix:proof_optimal} for proof.
\end{theorem}

Then, as can be seen in \cref{cor:moderate_exact}, \cref{eq:full_obj} guarantees moderate sharpening.
\begin{corollary}
\label{cor:moderate_exact}
Suppose $A(q,o_i)=A_+ \ge 0$ for $o_i \in \mathcal{O}^+$ and $A(q,o_j)=A_- \le 0$ for $o_j \in \mathcal{O}^-$, then \cref{eq:optimal_exact} yields moderate sharpening.
\end{corollary}
\begin{proof}
From \cref{eq:optimal_exact}, for $o_i \in \mathcal{O}^+$, $\pi^*(o_i|q) = \pi_{\text{ref}}(o_i|q)\,e^{A_+/\beta}\,/\,Z(q)$.
Since $A_+ \ge 0$, we have $e^{A_+/\beta} \ge 1$.
Moreover, $Z(q) = \sum_{o_i \in \mathcal{O}^+} \pi_{\text{ref}}(o_i|q) e^{A_+/\beta} + \sum_{o_j \in \mathcal{O}^-} \pi_{\text{ref}}(o_j|q) e^{A_-/\beta} \le e^{A_+/\beta}$ because $e^{A_-/\beta} \le 1$.
Thus $\pi^*(o_i|q) \ge \pi_{\text{ref}}(o_i|q)$ for $o_i \in \mathcal{O}^+$.
\end{proof}

\textbf{Finite‑sample sharpening behavior.}
In practice, we approximate \cref{eq:full_obj} using a finite batch of $G$ samples $\{o^{(1)}, \cdots, o^{(G)}\} \sim \pi_{\theta}(\cdot|q)$.
This yields an empirical objective where the gradient updates incrementally push the policy toward its stationary point.
We characterize this stationary point as the batch-optimal policy $\hat{\pi}$, which admits a closed-form solution analogous to the exact case but incorporates the empirical frequency $N_i$ of each mode in the batch.
Let $N_i$ denote the number of times mode $o_i$ appears in the batch ($\sum_i N_i = G$).
\cref{thm:batch_opt} gives the batch‑optimal policy $\hat{\pi}$.
\begin{theorem}
\label{thm:batch_opt}
The batch-optimal policy $\hat{\pi}$ is given by
\begin{equation}
\label{eq:batch_optimal}
\hat{\pi}(o_i|q) = \frac{1}{Z'(q)}\, \pi_{\text{ref}}(o_i|q)\,
\exp\!\left( \frac{N_i\, A(q,o_i)}{\beta G} \right),
\end{equation}
where $Z'(q) = \sum_j \pi_{\text{ref}}(o_j|q) \exp\!\bigl(N_j A(q,o_j)/(\beta G)\bigr)$.
See \cref{appendix_proof_batch_opt} for proof.
\end{theorem}
We emphasize that a single gradient step will generally not reach $\hat{\pi}$ exactly, but the update can be viewed as a geometric interpolation between the current policy and the batch-optimal target (see \cref{thm:grad_interp}), so the qualitative insights carry over.

To understand the sharpening dynamics of \cref{eq:batch_optimal}, we examine $\Delta_i = \hat{\pi}(o_i|q) - \pi_{\text{ref}}(o_i|q)$ in which $\sum_i \Delta_i = 0$.
We partition the modes ($\mathcal{O}^+ \cup \mathcal{O}^-$) into three categories based on empirical frequency $N_i$ and advantage sign $A(q, o_i)$.
Following the logic of \cref{cor:moderate_exact}, sampled modes ($N_i > 0$) reallocate mass according to advantage signs.
For any unsampled mode $o_i$ with $N_i = 0$, the exponential term in \cref{eq:batch_optimal} becomes 1.
Consequently, the update simplifies to a uniform scaling:
\begin{equation}
\nonumber
\hat{\pi}(o_i|q) = \frac{\pi_{\text{ref}}(o_i|q)}{Z'(q)} \implies
\Delta_i = \pi_{\text{ref}}(o_i|q)( \frac{1}{Z'(q)} - 1 ).
\end{equation}
If $Z'(q) > 1$, the probabilities of all unsampled modes decrease, and vice versa.
Since this set may contain valid reasoning paths, $Z'(q) > 1$ can induce over-sharpening.

\begin{theorem}
\label{thm:Z_bound}
(Proof in \cref{appendix_proof_thm3}.) 
Let $S^+$ and $S^-$ be the indices of sampled correct and incorrect modes.
Then the batch partition function $Z'(q)$ in \cref{thm:batch_opt} satisfies:
\begin{align}
Z'(q) \ge 1  + \frac{1}{\beta G}
&\big[
 A_+ \sum_{i \in S^+} N_i \,\pi_{\mathrm{ref}}(o_i \mid q) \nonumber \\
 &- |A_-| \sum_{j \in S^-} N_j \,\pi_{\mathrm{ref}}(o_j \mid q)
\big]. \nonumber
\end{align}

For normalized binary advantages $A_+ = (1 - p^+)/\sigma, A_- = -p^+/\sigma$ with batch accuracy $p^+$ and standard deviation $\sigma$, this bound simplifies to:
$Z'(q)
\;\ge\;
1 + \frac{p^+ (1 - p^+)}{\beta \sigma}\,\Delta_{\pi}.$
where $\Delta_{\pi} = \min_{i \in S^+} \pi_{\mathrm{ref}}(o_i | q) - \max_{j \in S^-} \pi_{\mathrm{ref}}(o_j | q)$.

\end{theorem}

As shown in \cref{thm:Z_bound}, the sharpening behavior is governed by the probability gap $\Delta_\pi$.
Whenever $\Delta_{\pi} > 0$, we see $Z'(q) > 1$ and therefore a uniform suppression of unsampled modes.
Intuitively, $\Delta_{\pi} > 0$ arises when the model is already performing well on the given query.
In fact, this also explains many existing methods (e.g., filtering out high‑confidence/easy samples~\cite{DAPO,sample_filtering,sample_filtering2,diffcult_filter}) as implicit ways of preventing $Z'(q)$ from becoming too large.
Furthermore, while the term $p^+(1-p^+)/\sigma$ stabilizes updates from skewed batches (vanishing as $p^+ \to \{0,1\}$), it is insufficient to counteract the intrinsic optimization bias that favors suppressing unsampled modes once model confidence is established.



\subsection{Semantic Coupling}
\label{sec2_3}

We now analyze how over‑sharpening propagates from a query $q$ to a positively semantically related query $q'$, e.g., two math problems that rely on the same underlying theorem.
This implies that a parameter update improving the model's performance on $q$ should also increase the likelihood of the correct response for $q'$.
To formalize this, we absorb the sign of the advantage $A(q, o)$ into the gradient direction and adopt \cref{assump:nonneg_grad}.

\begin{assumption}
\label{assump:nonneg_grad}
For positively semantically related pairs $(q, o)$ and $(q', o')$, their logit gradients are non-negatively aligned: $\nabla_\theta f(q, o)^\top \nabla_\theta f(q', o') \;\ge\; 0.$
\end{assumption}

The batch-optimal policy $\hat{\pi}$ prescribes target logit shifts $\mathbf{y} \in \mathbb{R}^G$, where $y_i = \frac{N_i |A(q,o_i)|}{\beta G}$\footnote{Due to the sign-absorption convention, the advantage term takes its absolute value here.}.
With a sufficiently small learning rate \cite{ntk}, fitting these targets yields the parameter update\footnote{For readers unfamiliar with the optimization dynamics in the linearized regime, we provide the necessary background in \cref{appendix_background}.}:
$$
\Delta\theta = \mathbf{J}_q^\top (\mathbf{J}_q \mathbf{J}_q^\top)^{-1} \mathbf{t} = \mathbf{J}_q^\top \mathbf{K}_{qq}^{-1} \mathbf{y},
$$
where $\mathbf{J}_q$ is the Jacobian of the logits with respect to parameters, and $\mathbf{K}_{qq} = \mathbf{J}_q \mathbf{J}_q^\top$ is the empirical kernel matrix on the current batch.
The update induces a logit shift at $(q', o')$:
\begin{align}
\Delta f(q', o') = \nabla_\theta f(q', o')^\top \Delta \theta = \mathbf{k}_{q'}^\top \mathbf{K}_{{qq}}^{-1} \mathbf{y} \nonumber,
\end{align}
where $\mathbf{k}_{q'}$ is the kernel vector with entries $[\mathbf{k}_{q'}]_i = k((q', o'), (q, o^{(i)})) \triangleq \nabla_\theta f(q', o')^\top \nabla_\theta f(q, o^{(i)})$.

\begin{figure*}[!th]
    \centering
    \begin{subfigure}[b]{0.3\textwidth}
        \centering
        \includegraphics[width=\textwidth]{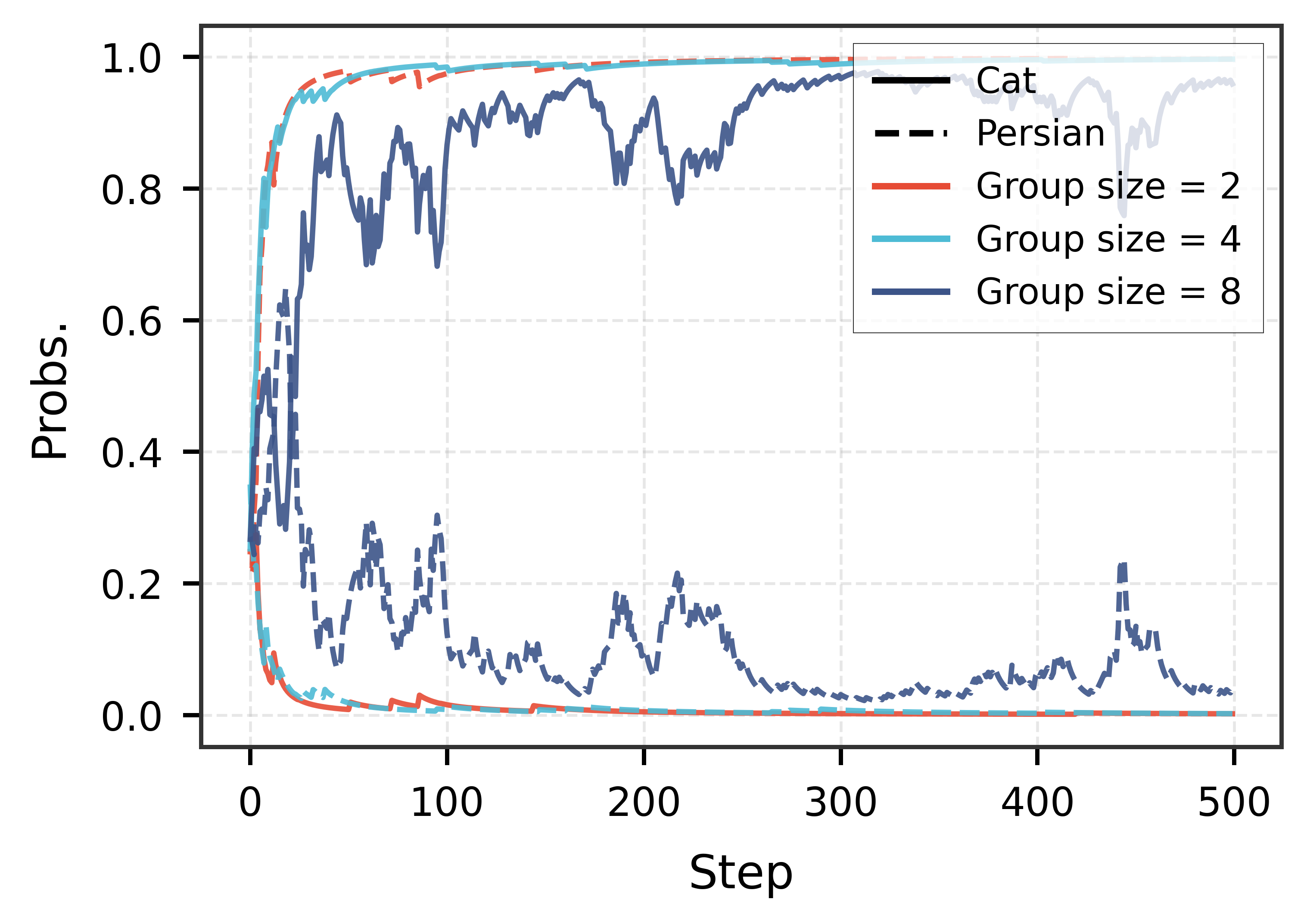}
    \end{subfigure}
    \hfill
    \begin{subfigure}[b]{0.3\textwidth}
        \centering
        \includegraphics[width=\textwidth]{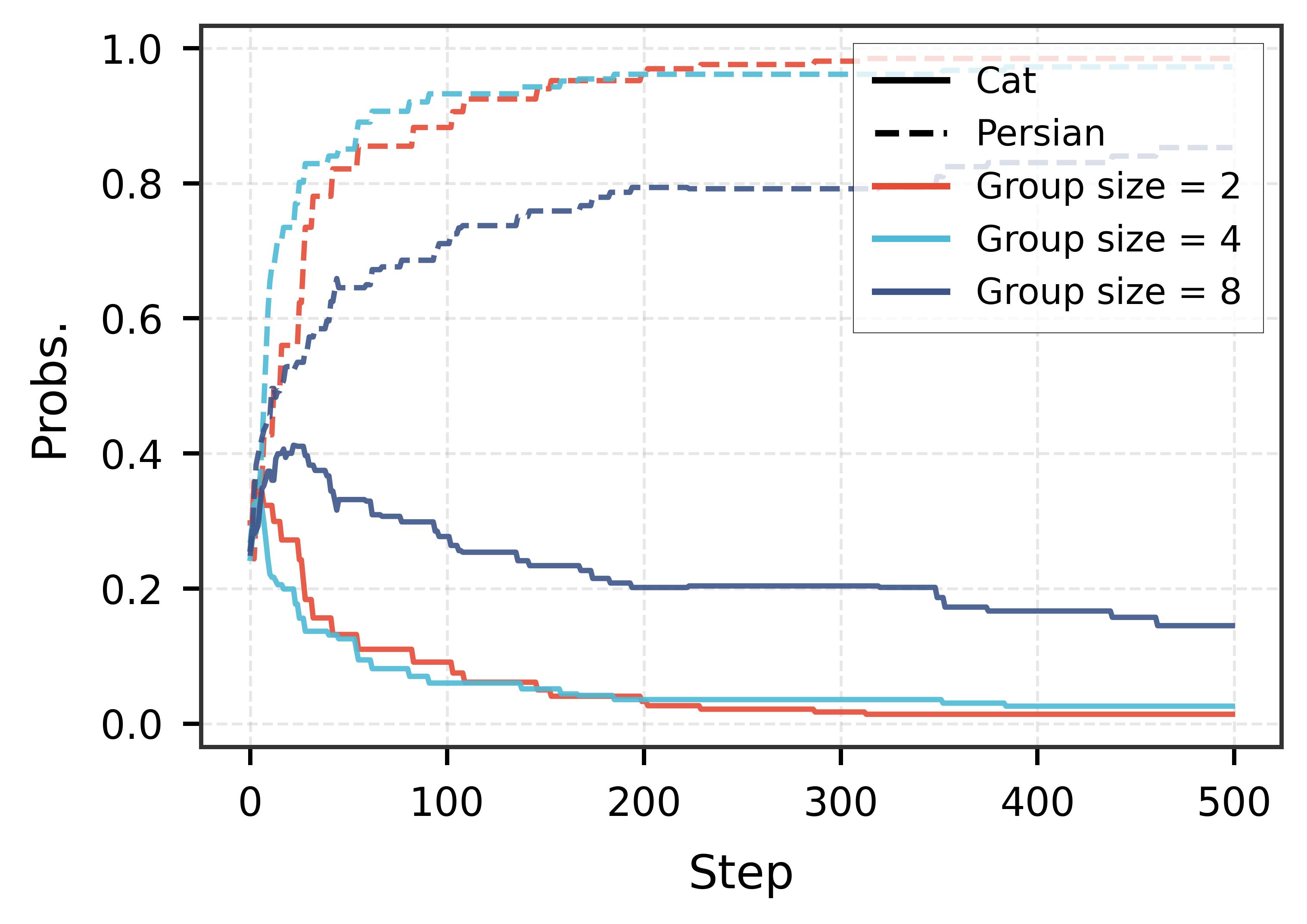}
    \end{subfigure}
    \hfill
    \begin{subfigure}[b]{0.3\textwidth}
        \centering
        \includegraphics[width=\textwidth]{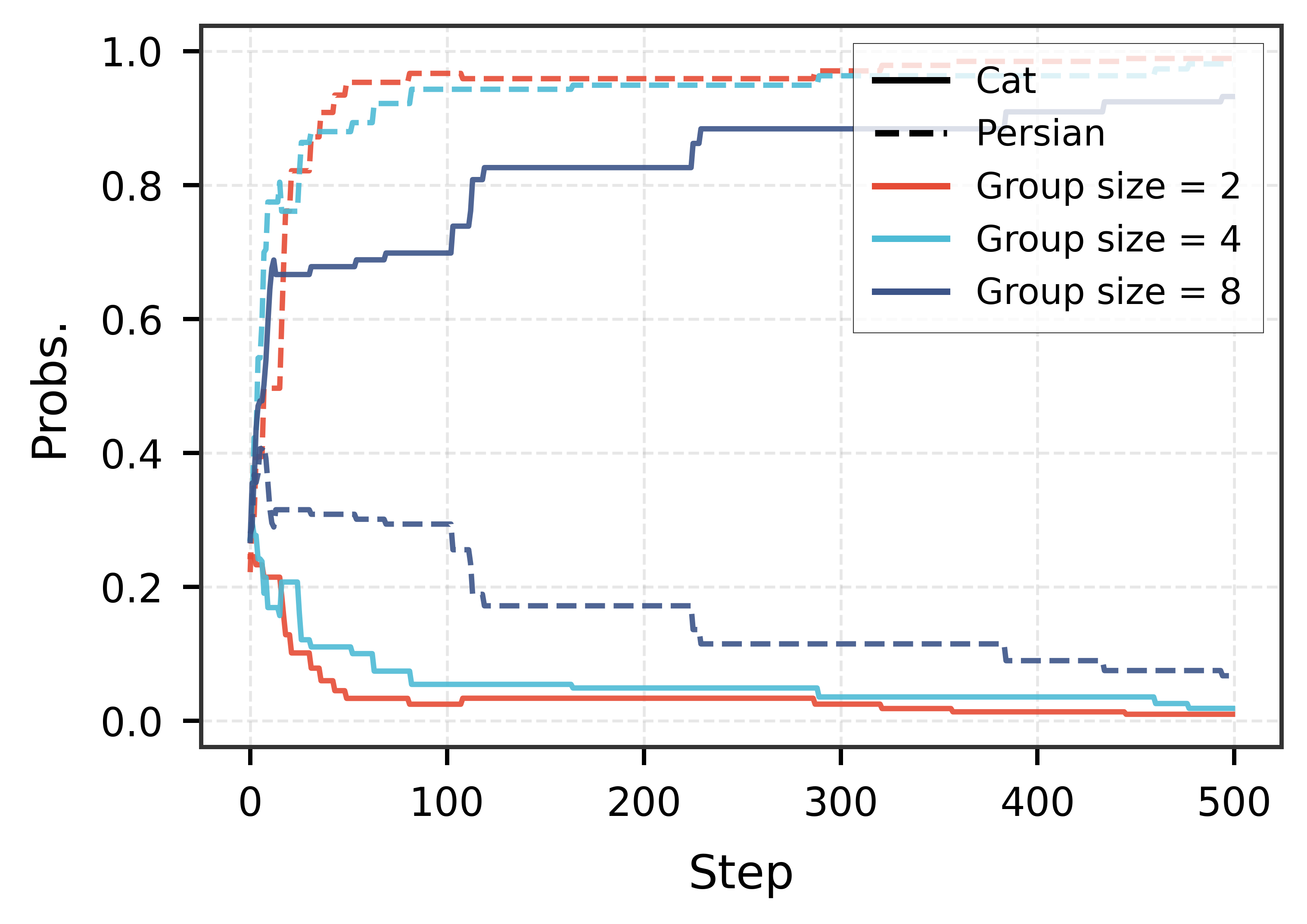}
    \end{subfigure}
    \caption{Effect of advantage estimation and group size on over-sharpening. Probability of Cat and Persian over training for raw (left), mean-shifted (mid), and normalized (right) advantages.}
    \label{fig:sampling_bias}
\end{figure*}

\begin{figure}[!th]
    \centering
    \centering
    \includegraphics[width=0.3\textwidth]{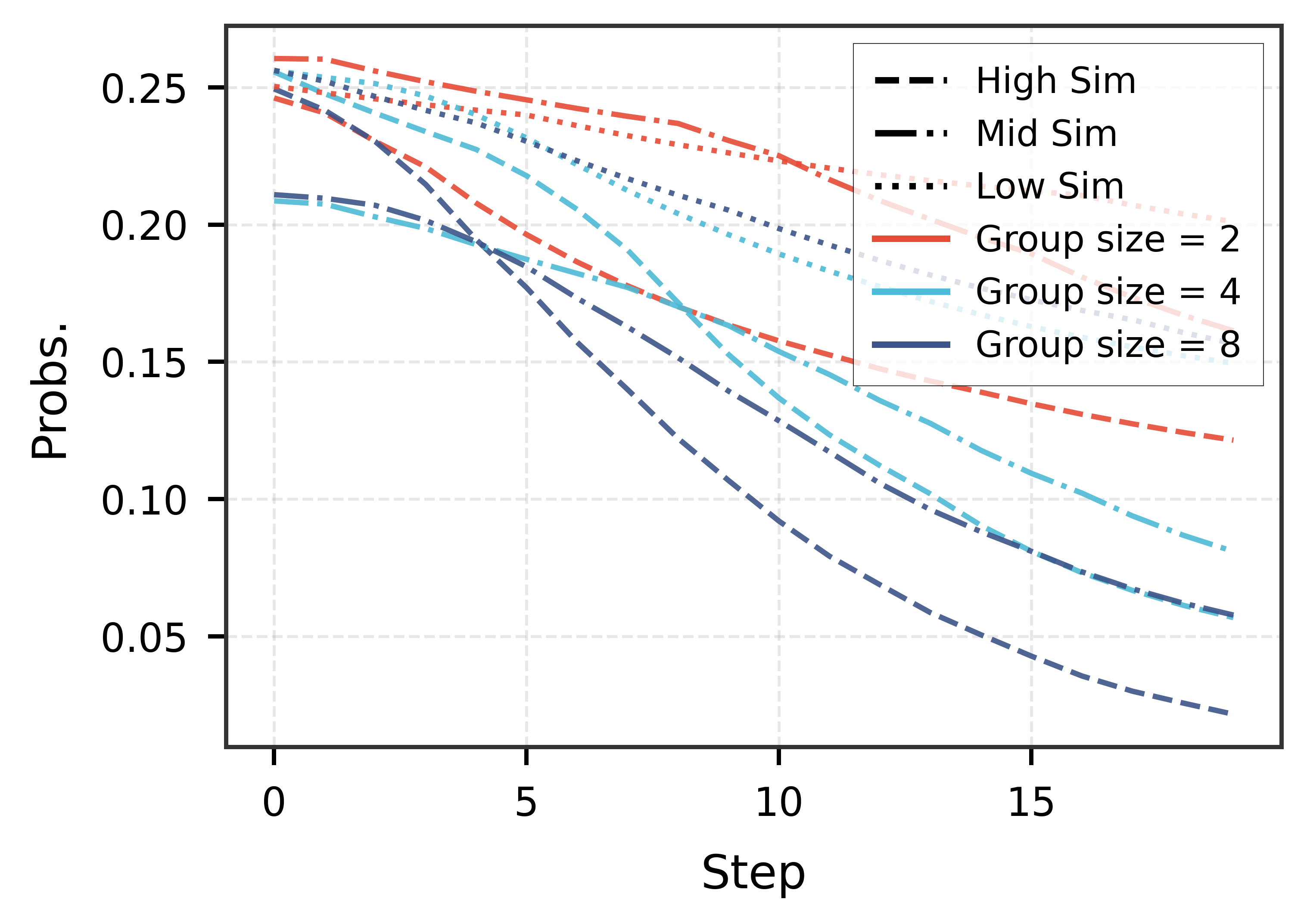}
    \caption{Semantic coupling accelerates collapse. Accuracy on the held-out Siamese class while training only on Persian. Stronger semantic coupling and larger group sizes both lead to faster degradation of Siamese accuracy.}
    \label{fig:semantic_coupling}
\end{figure}

Let $\mathbf{M}(\lambda, \rho) = (\lambda - \rho)\mathbf{I} + \rho \mathbf{1}\mathbf{1}^\top$, which possesses a principal eigenvalue $\lambda + (G-1)\rho$ associated with the constant vector $\mathbf{1}$, and a $(G-1)$-degenerate eigenvalue $\lambda - \rho$ for the remaining subspace.
Then, we can always find tight spectral bounds to encapsulate the actual spectrum of $\mathbf{K}_{qq}$: $0 \preceq \mathbf{M}(\lambda_{\min}, \rho_{\min}) \preceq \mathbf{K}_{qq} \preceq \mathbf{M}(\lambda_{\max}, \rho_{\max}).$
Here, $\lambda_{\min}$ and $\lambda_{\max}$ bound the diagonal entries (self‑alignment of each mode), while $\rho_{\min}$ and $\rho_{\max}$ bound the off‑diagonal entries (cross‑alignment between different modes).
In typical scenarios, updates meant for a particular mode influence that mode more than any other, which translates to $\rho_{\min} \leq \rho_{\max} < \lambda_{\min} \leq \lambda_{\max}$.
We formalize this as \cref{assump:diagonal_dominance}.
See \cref{appendix_ass_valid} for empirical validation of \cref{assump:nonneg_grad} and \cref{assump:diagonal_dominance}.
Then, to make the analysis manageable, we make \cref{assump:tranfer_kernel} that models the shift from a source query to a target query as a signal decay\footnote{\cref{assump:tranfer_kernel} can be relaxed to a bounded interval $[\eta_{\min}, \eta_{\max}]$ for greater generality, which would simply introduce a scaling factor $\frac{\eta_{\min}}{\eta_{\max}}$ to the ratio derived in \cref{corollary:ratio}.}.
Intuitively, $\eta$ can be interpreted as the similarity between $q'$ and $q$ from the model's perspective.
We then obtain the bound on the induced logit shifts, as shown in \cref{thm:logit_bounds}.

\begin{assumption}
\label{assump:tranfer_kernel}
Gradient correlations decay by a factor $\eta \in [0, 1]$ when shifting from query $q$ to $q'$, modeled as $k((q', o'), (q, o^{(s)})) \approx \eta K((q, o'), (q, o^{(s)}))$.
\end{assumption}

\begin{assumption}
\label{assump:diagonal_dominance}
Self-alignment consistently exceeds cross-alignment: $\rho_{\min} \leq \rho_{\max} < \lambda_{\min} \leq \lambda_{\max}.$
\end{assumption}

\begin{theorem}
\label{thm:logit_bounds}
Let $\mathcal{A}_{\text{sum}} \triangleq \sum_{i=1} y_i.$ 
The induced logit shift $\Delta f(q', o')$ satisfies the following bounds based on whether $o'$ appears in the training batch $\{o^{(i)}, i=1,\cdots,G\}$:
\begin{itemize}[leftmargin=*,topsep=1pt]
    \item  If $o' \notin \{o^{(i)}\}$, the logit increase is upper-bounded by:
    $$
    \Delta f(q', o') \leq  \frac{\eta \rho_{\max} \mathcal{A}_{\text{sum}}}{\lambda_{\min} + (G-1)\rho_{\min}}.
    $$
    \item If $o' = o^{(i)}$ for some $i \in \{1, \dots, G\}$, the logit increase is lower-bounded by
    \begin{equation}
    \begin{split}
    \nonumber
    &\Delta  f(q', o') = \Delta f(q', o_i) \geq\\ &\frac{\eta N_{o'} (\lambda_{\min} - \rho_{\min})}{\lambda_{\max} - \rho_{\max}} ( y_k - \frac{\rho_{\max} \mathcal{A}_{\text{sum}}}{\lambda_{\max} + (G-1)\rho_{\max}} ).
    \end{split}
    \end{equation}
\end{itemize}
See \cref{appendix_proof_thm4} for proof.
\end{theorem}

We leverage \cref{thm:logit_bounds} to characterize when over‑sharpening occurs at $q'$.
Consider a semantically coupled pair $(q,q')$ and suppose that at $q'$ there exists at least one mode $o'$ that appears in the batch (Case 2) and at least one mode $\bar{o}$ that does not (Case 1).
Over‑sharpening at $q'$ occurs when the logit increase for the batch‑seen mode dominates that for the batch‑unseen mode, because after normalization the latter becomes suppressed.
To quantify this, \cref{corollary:ratio} compare the guaranteed minimum increase for a seen mode (the lower bound of Case 2) to the maximum possible increase for an unseen mode (the upper bound of Case 1).

\begin{corollary}
\label{corollary:ratio}
Define the ratio $\mathcal{R}$ of the lower bound in Case 2 to the upper bound in Case 1:
\begin{equation}
\begin{split}
\nonumber
\mathcal{R} \triangleq \frac{\lambda_{\min} - \rho_{\min}}{\lambda_{\max} - \rho_{\max}} & [ \frac{N_{o'} y_k}{\mathcal{A}_{\text{sum}}} \cdot ( \frac{\lambda_{\min} + (G-1)\rho_{\min}}{\rho_{\max}} ) \\ &- \frac{\lambda_{\min} + (G-1)\rho_{\min}}{\lambda_{\max} + (G-1)\rho_{\max}} ].
\end{split}
\end{equation}
In particular, if $\lambda_{\min} \approx \lambda_{\max} \approx \lambda$ and $\rho_{\min} \approx \rho_{\max} \approx \rho$ (i.e., $\mathbf{K}_{qq}$ is nearly a diagonally dominant matrix with uniform diagonal and off‑diagonal entries), $\mathcal{R}$ simplifies to 
$\left[ \frac{N_{o'} y_k}{\mathcal{A}_{\text{sum}}} \left( \frac{\lambda}{\rho} + G - 1 \right) - 1 \right].
$
See \cref{appendix_proof_thm5} for proof.
\end{corollary}

The ratio $\mathcal{R}$ captures the relative strength with which a batch‑seen mode at $q'$ is reinforced compared to a semantically related but batch‑unseen mode.
A large $\mathcal{R}$ indicates that the guaranteed increase for the seen mode significantly outweighs the maximum possible increase for the unseen mode, leading to a disproportionate redistribution of probability mass and thus over‑sharpening at $q'$.
From the explicit form, over‑sharpening is promoted when:
\begin{itemize}[leftmargin=*,topsep=1pt]
    \item A large value of $N_{o'} y_i / \mathcal{A}_{\text{sum}}$ means that most of the update budget is focused on the particular mode that overlaps between $q$ and $q'$.
    \item Large $\lambda / \rho$ corresponds to highly specific features: gradient updates are tightly aligned with individual modes and leak only weakly to other modes.
    In this regime, improvement on $q$ reinforces a particular mode at $q'$ rather than broadly supporting all semantically consistent solutions.
    \item A larger effective batch size $G$ amplifies the effect.
\end{itemize}


\subsection{Empirical Illustration of Over‑Sharpening}
\label{sec2_4}



We use a controlled softmax classification setting to separately probe sampling bias and semantic coupling.
We consider two superclasses (Cat, Dog) and two subclasses (Persian, Siamese) with fixed feature embeddings:
\[
\mathbf{e}_{\text{Cat}} = [0.5, 0.5, 0.5, 0.1],\ 
\mathbf{e}_{\text{Persian}} = [0.75, 0.5, 0.25, 0.1],
\]
\[
\mathbf{e}_{\text{Dog}} = [0.1, 0.1, 0.1, 0.9], \ 
\mathbf{e}_{\text{Siamese}} = [0.25, 0.5, 0.75, 0.1].
\]
Persian and Siamese align with the Cat superclass but differ in subclass dimensions.
Treating inputs as $q$ and labels as $o$, we train this classifier on Persian inputs for 500 steps using SGD.
At each step, we sample $G$ outputs labels from the current softmax policy and assign reward 1 if the sampled label is Cat or Persian, and 0 otherwise, i.e., both the superclass and subclass are treated as correct for Persian.
We update the policy using various advantage estimators (raw, mean-shifted, normalized) and track the probabilities $\pi(o=\text{Cat} | q=\text{Persian})$ and $\pi(o=\text{Persian} | q=\text{Persian})$ for different group sizes $G$, as shown in \cref{fig:sampling_bias}.

Across all settings, the model eventually collapses onto a single label (either Cat or Persian, driven randomly by early sampling noise).
Once one mode is sampled more often, it dominates subsequent updates.
Increasing $G$ can delay collapse, but does not prevent this collapse.
The choice of the advantage estimator primarily modulates the rate of collapse.
Raw rewards are most aggressive, while normalized advantages are the slowest, aligning with our theory on $Z'(q)$ stabilization.
We also evaluate advanced advantage estimators, including RLOO and Reinforce++ with baseline, and observe that the collapse persists.
Moreover, momentum and adaptive optimizers (MI, AdamW) typically exacerbate over‑sharpening by accumulating past gradients and amplifying early sampling biases, pushing the policy more quickly into the collapsed regime.
See \cref{appendix_sup_exp_sec24} for these results.

We next study semantic coupling by monitoring the held-out Siamese subclass ($\pi(o=\text{Siamese}\mid q=\text{Siamese})$) while training on Persian.
Our theory shows that the propagation of collapse is governed by the interplay between the transfer intensity $\eta$ (\cref{thm:logit_bounds}), and the relative suppression ratio $\mathcal{R}$.
Notably, $\mathcal{R}$ scales with $G$.
To test this, we vary Siamese similarity (High, Mid, Low) by progressively shifting embedding mass from Persian-specific (dim 1) to Siamese-specific (dim 3) coordinates:
\[
[0.25, 0.5, 0.75, 0.1],
[0.1, 0.5, 0.9, 0.1],
[0.0, 0.5, 1.0, 0.1],
\]
As shown in \cref{fig:semantic_coupling}, Siamese accuracy is strongly squeezed during training.
This effect worsens with higher similarity, confirming that increased $\eta$ (high similarity) accelerates collapse.
Similarly, larger $G$ exacerbates over-sharpening.

\section{Approach}
\label{sec3}

We introduce two complementary mechanisms to keep RLVR in a moderate sharpening regime.
Inverse-success advantage calibration (IAC) adjusts how strongly each sampled trajectory contributes to the policy update.
Distribution-level calibration (DLC) tunes the rollout distribution to prevent the policy from repeatedly sampling the same high-probability patterns.
\cref{appendix_sup_exp_sec3} details the full pipeline (\cref{alg:rlvr_calibrated}) and visualizations demonstrating their effectiveness in mitigating over-sharpening via the Softmax case used in \cref{sec2_4}.

\subsection{Inverse‑success Advantage Calibration (IAC)}

\textbf{Idealized inverse‑probability reweighting.}
A natural solution would be to reweight advantages inversely proportional to $\pi_{\text{ref}}(\cdot)$, so that the reference‑probability factor in the bound cancels out.  
In detail, let $\tilde{A}_k = A(q,o_k) / \pi_{\text{ref}}(o_k)$.
Then the lower bound becomes:
\begin{align}
Z'(q) &\geq 1 + \frac{1}{\beta G} \sum_{k=1}^G N_k \tilde{A}_k \pi_{\text{ref}}(o_k) \nonumber \\
&= 1 + \frac{1}{\beta G} \sum_{k=1}^G N_k \left( \frac{A_k}{\pi_{\text{ref}}(o_k)} \right) \pi_{\text{ref}}(o_k) \nonumber  \\
&= 1 + \frac{1}{\beta G} \sum_{k=1}^G N_k A_k=1. \nonumber
\end{align}
In the last line, we use the normalization property $\sum_k N_k A_k = 0$.
However, this ideal solution is not directly usable in large‑scale RLVR for two reasons:
\begin{itemize}[leftmargin=*,topsep=1pt]
    \item \textbf{High variance in sequence probabilities.}  
   LLMs exhibit extremely high variance in sequence‑level probabilities.
   After calibration, this variance transfers to the advantages, which in turn causes highly unstable gradient updates.
   \item \textbf{Negative samples are not purely wrong.} 
   Negative samples often contain partially correct reasoning segments or are just being truncated due to a maximum length constraint.
   Over‑amplifying their (negative) advantages can therefore destroy useful intermediate structure and destabilizes training.
\end{itemize}

\textbf{Inverse-success advantage calibration.}
Motivated by the above, we retain the spirit of inverse reweighting but replace raw probabilities by a coarse, query‑level difficulty proxy.  
We only calibrate positive advantages and leave negative ones untouched.  
For each query $q$, let $S^+$ be the set of positive trajectories in the batch and $G$ the total number of trajectories.  
We define the calibrated advantages:
\begin{align}
    \tilde{A}(q, o_k^+) &= A(q, o_k^+) \cdot (G - |S^+|)^\alpha, \quad \alpha \ge 0, \\
    \tilde{A}(q, o_k^-) &= A(q, o_k^-),
\end{align}
where $o_k^+$ and $o_k^-$ denote positive and negative trajectories, respectively.
This simple heuristic enjoys two desirable properties:
This design has two key properties.
(1) $(G - |S^+|)$ serves as a difficulty proxy, assigning larger weights to harder queries where positive samples are rare.
(2) The exponent $\alpha$ governs the overall calibration strength. $\alpha=0$ recovers the baseline, while larger values de-emphasize easy queries.
Intuitively, IAC redistributes the advantage budget from easy queries to difficult ones, preventing excessive growth of $Z'(q)$ and mitigating over-sharpening.
By keeping negative advantages unchanged, we can avoid destabilizing training on partially correct or truncated trajectories.

\subsection{Distribution-Level Calibration via Decoupled Sampling (DLC)}

While IAC moderates how strongly each batch can inflate $Z'(q)$, it does not address that once certain trajectories have been suppressed and their probabilities become very small, they almost never appear again in subsequent rollouts.
To mitigate this, we suggest calibrating the sampling distribution itself.

A vanilla approach would be to increase the global sampling temperature, thereby flattening the policy distribution.
However, a global temperature treats all tokens equally, some modes are already well‑explored and saturated, while others remain under‑explored.
We instead propose a distribution-Level dalibration (DLC) based on a memory network that tracks the empirical rollout distribution.

DLC introduces an auxiliary memory network $m_\phi$ with logit function $f_\phi(q, o)$, intended to approximate the empirical frequency of $(q, o)$ pairs generated during training.
At each training step, we update $m_\phi$ on the sampled trajectories (with a standard cross‑entropy loss) so that it outputs higher logits for frequently generated outputs and lower logits for rarely seen ones.

During rollout, DLC constructs a calibrated sampling distribution by subtracting the memory logits from the policy logits as follows:
\begin{align}
    &\tilde{f}(q, o) = f_\theta(q, o) -  \mu \cdot f_\phi(q, o), \nonumber \\
    &\pi_{\text{sample}}(o \mid q) \propto \exp\big(\tilde{f}(q, o)\big). \nonumber
\end{align}
where $\mu \ge 0$ is a balance factor controlling how strongly the memory prior influences exploration.
Intuitively, the memory network acts as a learned frequency prior.
Wherein, modes that have been sampled many times receive larger $f_{\phi}(q,o)$ and are thus down‑weighted in $\tilde{f}(q,o)$, while under‑explored modes are relatively boosted.
From the perspective of \cref{thm:batch_opt} and \cref{thm:Z_bound}, by explicitly discouraging repeated sampling of the same high‑probability modes, it prevents any single mode from accumulating a disproportionately large empirical count $N_i$ across updates.

\textbf{Computational overhead and practical choice.}
Compared to IAC, DLC introduces non‑trivial computational overhead because it requires both policy and memory forward passes during rollout, and training updates for $m_\phi$.  
If $m_\phi$ shares the same architecture as the policy, the theoretical cost roughly doubles. 
In our implementation, the dominant additional cost comes from rollout.
Although integrated with high-performance rollout backends like vLLM, the lack of kernel‑level support for synchronized multi‑model sampling leads to approximately $5\times$ higher rollout latency.
Given this overhead, our default configuration in large‑scale experiments employs IAC alone, which is compute‑efficient and already substantially mitigates over‑sharpening.  
We treat DLC as an optional augmentation that can be enabled when compute permits.
Empirically, adding DLC on top of IAC yields further improvements, especially in regimes where over‑exploitation of a few modes is particularly pronounced.

\begin{table*}[!ht]
\caption{Comparison of AVG@8 and PASS@8 performance across six benchmark datasets. } 
\label{table:merged_horizontal}
\centering
\setlength{\tabcolsep}{3pt}
\resizebox{\textwidth}{!}{%
\begin{tabular}{@{}cc|ccccccc|ccccccc@{}}
\toprule
\multirow{2}{*}{Model} & \multirow{2}{*}{Method} & \multicolumn{7}{c|}{AVG@8 (\%)} & \multicolumn{7}{c}{PASS@8 (\%)} \\ \cmidrule(l){3-9} \cmidrule(l){10-16}
 &  & GSM8K & Math500 & LightEval & AIME24 & Minerva & O-Bench & Avg. & GSM8K & Math500 & LightEval & AIME24 & Minerva & O-Bench & Avg. \\ \midrule
\multirow{5}{*}{Qwen3-4B} 
 & GRPO & 93.72 & 85.58 & \underline{70.61} & 34.58 & 43.06 & \underline{48.85} & 62.73 & 96.74 & 91.94 & \underline{75.43} & 47.37 & 54.54 & \underline{56.44} & 70.41 \\
 & RLOO & 93.69 & \textbf{86.44} & 70.24 & 32.50 & 43.98 & 47.91 & 62.46 & \underline{96.98} & \underline{92.22} & 75.39 & 43.65 & 55.01 & 56.31 & 69.93 \\
 & Reinforce++ & \underline{94.04} & 85.43 & 70.21 & \underline{36.25} & \underline{45.63} & 48.57 & \underline{63.36} & 96.45 & 91.58 & 74.92 & \underline{50.54} & \underline{55.32} & 56.38 & \underline{70.87} \\
 & DrGRPO & 94.10 & 83.56 & 69.54 & 35.42 & 45.43 & 47.83 & 62.65 & 96.43 & 89.90 & 73.87 & 47.02 & 54.60 & 55.08 & 69.48 \\
 & IAC (Ours) & \textbf{94.63} & \underline{86.43} & \textbf{72.15} & \textbf{37.08} & \textbf{47.01} & \textbf{48.87} & \textbf{64.36} & \textbf{97.38} & \textbf{92.33} & \textbf{77.30} & \textbf{52.28} & \textbf{57.74} & \textbf{56.59} & \textbf{72.27} \\ \midrule
\multirow{5}{*}{Qwen3-8B} 
 & GRPO & 95.59 & 85.81 & 71.12 & 35.42 & 47.20 & 49.79 & 64.16 & 96.51 & 91.39 & 75.19 & 52.36 & 55.65 & 56.88 & 71.33 \\
 & RLOO & 95.56 & 86.89 & 72.30 & 36.25 & 46.78 & 49.74 & 64.59 & \underline{97.25} & 92.64 & 76.17 & 50.92 & 55.55 & 57.68 & 71.70 \\
 & Reinforce++ & \underline{95.56} & \underline{87.07} & \underline{72.14} & \textbf{40.68} & \underline{47.84} & \underline{50.30} & \underline{65.60} & 96.96 & 92.31 & 75.96 & \textbf{55.24} & 56.12 & \underline{57.78} & \underline{72.40} \\
 & DrGRPO & 95.35 & 86.11 & 71.56 & 34.58 & 47.38 & 49.87 & 64.14 & 97.11 & 91.57 & 75.36 & 52.25 & 55.41 & 57.26 & 71.49 \\
 & IAC (Ours) & \textbf{95.72} & \textbf{87.67} & \textbf{73.85} & 35.42 & \textbf{50.60} & \textbf{50.49} & \textbf{65.63} & \textbf{97.56} & \textbf{94.54} & \textbf{78.69} & \underline{52.67} & \textbf{60.60} & \textbf{58.48} & \textbf{73.76} \\ \midrule
\multirow{5}{*}{Qwen3-14B} 
 & GRPO & 95.96 & 89.42 & 73.83 & 53.45 & 50.32 & 55.28 & 69.71 & 97.53 & 93.75 & 76.54 & 70.44 & 59.40 & 60.87 & 76.42 \\
 & RLOO & \underline{96.04} & 90.29 & 73.64 & 54.17 & 50.46 & 55.10 & 69.95 & 97.43 & 93.62 & 76.36 & \underline{73.18} & 59.78 & 61.14 & 76.92 \\
 & Reinforce++ & 95.82 & \textbf{90.88} & \underline{73.96} & 54.17 & \underline{50.87} & \underline{56.00} & 70.28 & \underline{97.79} & 94.12 & \underline{76.74} & 71.71 & 60.27 & \underline{61.42} & \underline{77.01} \\
 & DrGRPO & 95.96 & 90.37 & 73.85 & \underline{54.85} & 50.69 & 55.35 & \underline{70.18} & 97.53 & \underline{94.27} & 76.57 & 71.83 & \underline{61.03} & 60.96 & 77.03 \\
 & IAC (Ours) & \textbf{96.36} & \underline{90.70} & \textbf{74.52} & \textbf{56.67} & \textbf{52.34} & \textbf{56.32} & \textbf{71.15} & \textbf{98.11} & \textbf{94.37} & \textbf{77.28} & \textbf{78.07} & \textbf{62.64} & \textbf{62.37} & \textbf{78.81} \\ \bottomrule
\end{tabular}%
}
\end{table*}

\begin{table}[!ht]
\centering
\caption{Compatibility analysis using Qwen3-8B.}
\label{table:compatibility}
\scriptsize
\begin{tabular}{@{}cccccc@{}}
\toprule
Metric                 & Method    & Math500 & LightEval & Minerva & O-Bench \\ \midrule
\multirow{4}{*}{AVG@8}  & DAPO      & 87.25   & 72.04     & 47.15   & 50.17   \\
                       & IAC      & 87.67   & 73.85     & 50.60   & 50.49   \\
                       & DAPO+IAC & 87.89   & 74.48     & 51.07   & 50.58   \\
                       & DAPO+IAC+DLC       & \textbf{88.44}   & \textbf{75.42}     & \textbf{51.90}   & \textbf{51.38}   \\ \midrule \midrule
\multirow{4}{*}{PASS@8} & DAPO      & 92.96   & 76.62     & 56.38   & 58.28   \\
                       & IAC      & 94.54   & 78.69     & 60.60   & 58.48   \\
                       & DAPO+IAC & 94.74   & 84.80     & 61.97   & 58.88   \\
                       & DAPO+IAC+DLC       & \textbf{95.19}   & \textbf{85.15}     & \textbf{62.50}   & \textbf{60.12}   \\ \bottomrule
\end{tabular}
\end{table}

\section{Empirical Evaluation}
\label{sec4}

\subsection{Experimental Setup}

\textbf{Models, datasets, and metrics.}
We evaluate our methods in mathematical reasoning with verifiable rewards.
We use the DeepScaleR~\cite{deepscaler} corpus as our primary RLVR training set, consisting of approximately 40.3k problem–answer pairs. 
To assess generalization, we report performance on six diverse benchmarks: GSM8K, AIME 2024, MATH 500, LightEval, Minerva~\cite{Minerva}, and O‑Bench~\cite{OlympiadBench}.
We conduct experiments with three backbone models, Qwen3‑4B, Qwen3-8B and Qwen3‑14B~\cite{qwen3}.
As evaluation metrics, we report PASS@8 and AVG@8.
PASS@8 measures the proportion of queries for which at least one of the eight responses is correct, and AVG@8 reflects the average proportion of correct responses per query.
\cref{appendix_sup_exp_sec4} provides evaluation results on different models and datasets.

\textbf{Baselines.}
We compare our approaches against a range of RL baselines commonly used in LLM training, including GRPO~\cite{deepseek_r1}, DrGRPO~\cite{DrGRPO}, RLOO~\cite{RLOO}, Reinforce++ with baseline~\cite{reinforce++}, and DAPO~\cite{DAPO}.
We also compare IAC with sample-filtering-based methods~\cite{diffcult_filter} in \cref{appendix_sup_exp_sec4}.
All baselines are run in the same setting with identical reward functions and sampling budgets for fair comparison.

\textbf{Implementation details.}
All models (including the memory network) are optimized with AdamW using a learning rate of $10^{-6}$.
We employ a global batch size of 64 and sample $G=$ 8 responses per query.
Training is run for 2000 steps, and we evaluate checkpoints every 50 steps, reporting the peak performance for each method.
We employ VERL’s built-in recipes to standardize the implementation of baseline methods.
For IAC, we use $\alpha=1$ by default.
Due to the page limit, we leave sensitivity analysis ($\alpha$), the comparison with sample-filtering baselines, and the entropy evolution analysis in \cref{appendix_sup_exp_sec4}.
All experiments are conducted using 32 H200 GPUs.

\subsection{Evaluation Result}

\textbf{Comparison with state-of-the-art methods.}
\cref{table:merged_horizontal} summarizes the AVG@8 and PASS@8 performance of different methods, respectively, across six mathematical reasoning benchmarks and three Qwen3 backbones.
Overall, our method attains the best AVG@8 and PASS@8 in almost all settings across models and datasets, indicating that calibrated sharpening consistently improves both average quality and success probability.
In contrast, different baselines occasionally excel on specific benchmarks.
For example, Reinforce++ is competitive on AIME24 and MATH500, and RLOO sometimes attains strong averages.
To illustrate this in more detail, consider the Qwen3‑14B results.
Here, all baselines are already strong, yet our method still delivers clear gains.
In terms of AVG@8, our method can raise the overall average from roughly 70.2 (Reinforce++ / DrGRPO) to 71.15.
For PASS@8, the improvement is more striking: from 77.03 (best baseline) to 78.81.
In particular, improvements on harder, less saturated benchmarks (AIME24, Minerva, O‑Bench) are more pronounced than on easier ones such as GSM8K, suggesting that our approach is especially beneficial when reasoning requires exploring multiple reasoning modes rather than amplifying a single dominant pattern.
On the smaller backbones Qwen3‑4B and Qwen3‑8B, we observe a consistent pattern.

\textbf{Compatibility with other methods.}
We also evaluate the compatibility of the proposed methods.
The results, presented in \cref{table:compatibility}, reveal three key insights.
First, IAC alone outperforms DAPO on both AVG@8 and PASS@8 metrics across most datasets.
Second, when IAC is integrated into DAPO (DAPO+IAC), we observe additive performance gains.
For instance, on LightEval (PASS@8), the combination improves performance from $76.62\%$ (DAPO) to $84.80\%$.
Third, we incorporate DLC with a memory network (Qwen3-1.7B) of $\mu=0.5$ (DAPO+IAC+DLC).
This full configuration yields the highest performance.

\section{Conclusion}
\label{sec6}

In this work, we formalized the phenomenon of over-sharpening within the RLVR framework.
Our theoretical analysis identifies two fundamental drivers of distribution collapse: finite-sample bias, which causes the uniform suppression of unsampled correct modes, and semantic coupling, which allows this collapse to propagate across related tasks through shared model parameters.
To mitigate these issues, we introduced two plug-and-play calibration mechanisms IAC and DLC that stabilize the partition function and diversify the sampling process.
Extensive evaluations across multiple reasoning benchmarks and model scales demonstrate that our methods effectively achieve consistent performance gains over state-of-the-art baselines.

\nocite{langley00}

\bibliography{icml_refs}

@article{deepseek_r1,
  author       = {DeepSeek{-}AI},
  title        = {DeepSeek-R1: Incentivizing Reasoning Capability in LLMs via Reinforcement
                  Learning},
  journal      = {CoRR},
  volume       = {abs/2501.12948},
  year         = {2025},
  url          = {https://doi.org/10.48550/arXiv.2501.12948},
  doi          = {10.48550/ARXIV.2501.12948},
  eprinttype    = {arXiv},
  eprint       = {2501.12948},
  bibsource    = {dblp computer science bibliography, https://dblp.org}
}

@article{uncertain_reward,
  author       = {Andre He and
                  Daniel Fried and
                  Sean Welleck},
  title        = {Rewarding the Unlikely: Lifting {GRPO} Beyond Distribution Sharpening},
  journal      = {CoRR},
  volume       = {abs/2506.02355},
  year         = {2025},
  url          = {https://doi.org/10.48550/arXiv.2506.02355},
  doi          = {10.48550/ARXIV.2506.02355},
  eprinttype    = {arXiv},
  eprint       = {2506.02355},
  bibsource    = {dblp computer science bibliography, https://dblp.org}
}

@article{reasoning_inference,
  author       = {Aayush Karan and
                  Yilun Du},
  title        = {Reasoning with Sampling: Your Base Model is Smarter Than You Think},
  journal      = {CoRR},
  volume       = {abs/2510.14901},
  year         = {2025},
  url          = {https://doi.org/10.48550/arXiv.2510.14901},
  doi          = {10.48550/ARXIV.2510.14901},
  eprinttype    = {arXiv},
  eprint       = {2510.14901},
  bibsource    = {dblp computer science bibliography, https://dblp.org}
}

@article{passk_better_base_model,
  author       = {Yang Yue and
                  Zhiqi Chen and
                  Rui Lu and
                  Andrew Zhao and
                  Zhaokai Wang and
                  Yang Yue and
                  Shiji Song and
                  Gao Huang},
  title        = {Does Reinforcement Learning Really Incentivize Reasoning Capacity
                  in LLMs Beyond the Base Model?},
  journal      = {CoRR},
  volume       = {abs/2504.13837},
  year         = {2025},
  url          = {https://doi.org/10.48550/arXiv.2504.13837},
  doi          = {10.48550/ARXIV.2504.13837},
  eprinttype    = {arXiv},
  eprint       = {2504.13837},
  bibsource    = {dblp computer science bibliography, https://dblp.org}
}

@article{passk_better_base_model2,
  author       = {Yuda Song and
                  Julia Kempe and
                  R{\'{e}}mi Munos},
  title        = {Outcome-based Exploration for {LLM} Reasoning},
  journal      = {CoRR},
  volume       = {abs/2509.06941},
  year         = {2025},
  url          = {https://doi.org/10.48550/arXiv.2509.06941},
  doi          = {10.48550/ARXIV.2509.06941},
  eprinttype    = {arXiv},
  eprint       = {2509.06941},
  bibsource    = {dblp computer science bibliography, https://dblp.org}
}

@article{minimize_entropy,
  author       = {Mihir Prabhudesai and
                  Lili Chen and
                  Alex Ippoliti and
                  Katerina Fragkiadaki and
                  Hao Liu and
                  Deepak Pathak},
  title        = {Maximizing Confidence Alone Improves Reasoning},
  journal      = {CoRR},
  volume       = {abs/2505.22660},
  year         = {2025},
  url          = {https://doi.org/10.48550/arXiv.2505.22660},
  doi          = {10.48550/ARXIV.2505.22660},
  eprinttype    = {arXiv},
  eprint       = {2505.22660},
  bibsource    = {dblp computer science bibliography, https://dblp.org}
}

@article{entropy_reward,
  author       = {Daixuan Cheng and
                  Shaohan Huang and
                  Xuekai Zhu and
                  Bo Dai and
                  Wayne Xin Zhao and
                  Zhenliang Zhang and
                  Furu Wei},
  title        = {Reasoning with Exploration: An Entropy Perspective},
  journal      = {CoRR},
  volume       = {abs/2506.14758},
  year         = {2025},
  url          = {https://doi.org/10.48550/arXiv.2506.14758},
  doi          = {10.48550/ARXIV.2506.14758},
  eprinttype    = {arXiv},
  eprint       = {2506.14758},
  bibsource    = {dblp computer science bibliography, https://dblp.org}
}

@article{qwen3,
  author       = {An Yang and
                  Anfeng Li and
                  Baosong Yang and
                  Beichen Zhang and
                  Binyuan Hui and
                  Bo Zheng and
                  Bowen Yu and
                  Chang Gao and
                  Chengen Huang and
                  Chenxu Lv and
                  Chujie Zheng and
                  Dayiheng Liu and
                  Fan Zhou and
                  Fei Huang and
                  Feng Hu and
                  Hao Ge and
                  Haoran Wei and
                  Huan Lin and
                  Jialong Tang and
                  Jian Yang and
                  Jianhong Tu and
                  Jianwei Zhang and
                  Jian Yang and
                  Jiaxi Yang and
                  Jingren Zhou and
                  Junyang Lin and
                  Kai Dang and
                  Keqin Bao and
                  Kexin Yang and
                  Le Yu and
                  Lianghao Deng and
                  Mei Li and
                  Mingfeng Xue and
                  Mingze Li and
                  Pei Zhang and
                  Peng Wang and
                  Qin Zhu and
                  Rui Men and
                  Ruize Gao and
                  Shixuan Liu and
                  Shuang Luo and
                  Tianhao Li and
                  Tianyi Tang and
                  Wenbiao Yin and
                  Xingzhang Ren and
                  Xinyu Wang and
                  Xinyu Zhang and
                  Xuancheng Ren and
                  Yang Fan and
                  Yang Su and
                  Yichang Zhang and
                  Yinger Zhang and
                  Yu Wan and
                  Yuqiong Liu and
                  Zekun Wang and
                  Zeyu Cui and
                  Zhenru Zhang and
                  Zhipeng Zhou and
                  Zihan Qiu},
  title        = {Qwen3 Technical Report},
  journal      = {CoRR},
  volume       = {abs/2505.09388},
  year         = {2025},
  url          = {https://doi.org/10.48550/arXiv.2505.09388},
  doi          = {10.48550/ARXIV.2505.09388},
  eprinttype    = {arXiv},
  eprint       = {2505.09388},
  bibsource    = {dblp computer science bibliography, https://dblp.org}
}

@article{rule_2080,
  author       = {Shenzhi Wang and
                  Le Yu and
                  Chang Gao and
                  Chujie Zheng and
                  Shixuan Liu and
                  Rui Lu and
                  Kai Dang and
                  Xionghui Chen and
                  Jianxin Yang and
                  Zhenru Zhang and
                  Yuqiong Liu and
                  An Yang and
                  Andrew Zhao and
                  Yang Yue and
                  Shiji Song and
                  Bowen Yu and
                  Gao Huang and
                  Junyang Lin},
  title        = {Beyond the 80/20 Rule: High-Entropy Minority Tokens Drive Effective
                  Reinforcement Learning for {LLM} Reasoning},
  journal      = {CoRR},
  volume       = {abs/2506.01939},
  year         = {2025},
  url          = {https://doi.org/10.48550/arXiv.2506.01939},
  doi          = {10.48550/ARXIV.2506.01939},
  eprinttype    = {arXiv},
  eprint       = {2506.01939},
  bibsource    = {dblp computer science bibliography, https://dblp.org}
}

@article{DAPO,
  author       = {Qiying Yu and
                  Zheng Zhang and
                  Ruofei Zhu and
                  Yufeng Yuan and
                  Xiaochen Zuo and
                  Yu Yue and
                  Tiantian Fan and
                  Gaohong Liu and
                  Lingjun Liu and
                  Xin Liu and
                  Haibin Lin and
                  Zhiqi Lin and
                  Bole Ma and
                  Guangming Sheng and
                  Yuxuan Tong and
                  Chi Zhang and
                  Mofan Zhang and
                  Wang Zhang and
                  Hang Zhu and
                  Jinhua Zhu and
                  Jiaze Chen and
                  Jiangjie Chen and
                  Chengyi Wang and
                  Hongli Yu and
                  Weinan Dai and
                  Yuxuan Song and
                  Xiangpeng Wei and
                  Hao Zhou and
                  Jingjing Liu and
                  Wei{-}Ying Ma and
                  Ya{-}Qin Zhang and
                  Lin Yan and
                  Mu Qiao and
                  Yonghui Wu and
                  Mingxuan Wang},
  title        = {{DAPO:} An Open-Source {LLM} Reinforcement Learning System at Scale},
  journal      = {CoRR},
  volume       = {abs/2503.14476},
  year         = {2025},
  url          = {https://doi.org/10.48550/arXiv.2503.14476},
  doi          = {10.48550/ARXIV.2503.14476},
  eprinttype    = {arXiv},
  eprint       = {2503.14476},
  bibsource    = {dblp computer science bibliography, https://dblp.org}
}

@article{cov_update,
  author       = {Ganqu Cui and
                  Yuchen Zhang and
                  Jiacheng Chen and
                  Lifan Yuan and
                  Zhi Wang and
                  Yuxin Zuo and
                  Haozhan Li and
                  Yuchen Fan and
                  Huayu Chen and
                  Weize Chen and
                  Zhiyuan Liu and
                  Hao Peng and
                  Lei Bai and
                  Wanli Ouyang and
                  Yu Cheng and
                  Bowen Zhou and
                  Ning Ding},
  title        = {The Entropy Mechanism of Reinforcement Learning for Reasoning Language
                  Models},
  journal      = {CoRR},
  volume       = {abs/2505.22617},
  year         = {2025},
  url          = {https://doi.org/10.48550/arXiv.2505.22617},
  doi          = {10.48550/ARXIV.2505.22617},
  eprinttype    = {arXiv},
  eprint       = {2505.22617},
  bibsource    = {dblp computer science bibliography, https://dblp.org}
}

@article{reinforce++,
  title={REINFORCE++: Stabilizing Critic-Free Policy Optimization with Global Normalization},
  author={Hu, Jian and Liu, Jason Klein and Xu, Haotian and Shen, Wei},
  journal={arXiv preprint arXiv},
  volume={2501}
}

@inproceedings{RLOO,
  author       = {Wouter Kool and
                  Herke van Hoof and
                  Max Welling},
  title        = {Attention, Learn to Solve Routing Problems!},
  booktitle    = {7th International Conference on Learning Representations, {ICLR} 2019,
                  New Orleans, LA, USA, May 6-9, 2019},
  publisher    = {OpenReview.net},
  year         = {2019},
  url          = {https://openreview.net/forum?id=ByxBFsRqYm},
  bibsource    = {dblp computer science bibliography, https://dblp.org}
}

@article{DrGRPO,
  author       = {Zichen Liu and
                  Changyu Chen and
                  Wenjun Li and
                  Penghui Qi and
                  Tianyu Pang and
                  Chao Du and
                  Wee Sun Lee and
                  Min Lin},
  title        = {Understanding R1-Zero-Like Training: {A} Critical Perspective},
  journal      = {CoRR},
  volume       = {abs/2503.20783},
  year         = {2025},
  url          = {https://doi.org/10.48550/arXiv.2503.20783},
  doi          = {10.48550/ARXIV.2503.20783},
  eprinttype    = {arXiv},
  eprint       = {2503.20783},
  bibsource    = {dblp computer science bibliography, https://dblp.org}
}

@article{deepscaler,
  title={Deepscaler: Surpassing o1-preview with a 1.5 b model by scaling rl},
  author={Luo, Michael and Tan, Sijun and Wong, Justin and Shi, Xiaoxiang and Tang, William Y and Roongta, Manan and Cai, Colin and Luo, Jeffrey and Zhang, Tianjun and Li, Li Erran and others},
  journal={Notion Blog},
  year={2025}
}

@inproceedings{OlympiadBench,
  author       = {Chaoqun He and
                  Renjie Luo and
                  Yuzhuo Bai and
                  Shengding Hu and
                  Zhen Leng Thai and
                  Junhao Shen and
                  Jinyi Hu and
                  Xu Han and
                  Yujie Huang and
                  Yuxiang Zhang and
                  Jie Liu and
                  Lei Qi and
                  Zhiyuan Liu and
                  Maosong Sun},
  editor       = {Lun{-}Wei Ku and
                  Andre Martins and
                  Vivek Srikumar},
  title        = {OlympiadBench: {A} Challenging Benchmark for Promoting {AGI} with
                  Olympiad-Level Bilingual Multimodal Scientific Problems},
  booktitle    = {Proceedings of the 62nd Annual Meeting of the Association for Computational
                  Linguistics (Volume 1: Long Papers), {ACL} 2024, Bangkok, Thailand,
                  August 11-16, 2024},
  pages        = {3828--3850},
  publisher    = {Association for Computational Linguistics},
  year         = {2024},
  url          = {https://doi.org/10.18653/v1/2024.acl-long.211},
  doi          = {10.18653/V1/2024.ACL-LONG.211},
  bibsource    = {dblp computer science bibliography, https://dblp.org}
}

@inproceedings{Minerva,
  author       = {Aitor Lewkowycz and
                  Anders Andreassen and
                  David Dohan and
                  Ethan Dyer and
                  Henryk Michalewski and
                  Vinay V. Ramasesh and
                  Ambrose Slone and
                  Cem Anil and
                  Imanol Schlag and
                  Theo Gutman{-}Solo and
                  Yuhuai Wu and
                  Behnam Neyshabur and
                  Guy Gur{-}Ari and
                  Vedant Misra},
  editor       = {Sanmi Koyejo and
                  S. Mohamed and
                  A. Agarwal and
                  Danielle Belgrave and
                  K. Cho and
                  A. Oh},
  title        = {Solving Quantitative Reasoning Problems with Language Models},
  booktitle    = {Advances in Neural Information Processing Systems 35: Annual Conference
                  on Neural Information Processing Systems 2022, NeurIPS 2022, New Orleans,
                  LA, USA, November 28 - December 9, 2022},
  year         = {2022},
  url          = {http://papers.nips.cc/paper\_files/paper/2022/hash/18abbeef8cfe9203fdf9053c9c4fe191-Abstract-Conference.html},
  bibsource    = {dblp computer science bibliography, https://dblp.org}
}

@article{qae,
  author       = {Junkang Wu and
                  Kexin Huang and
                  Jiancan Wu and
                  An Zhang and
                  Xiang Wang and
                  Xiangnan He},
  title        = {Quantile Advantage Estimation for Entropy-Safe Reasoning},
  journal      = {CoRR},
  volume       = {abs/2509.22611},
  year         = {2025},
  url          = {https://doi.org/10.48550/arXiv.2509.22611},
  doi          = {10.48550/ARXIV.2509.22611},
  eprinttype    = {arXiv},
  eprint       = {2509.22611},
  bibsource    = {dblp computer science bibliography, https://dblp.org}
}

@article{sample_filtering,
  author       = {Wei Xiong and
                  Jiarui Yao and
                  Yuhui Xu and
                  Bo Pang and
                  Lei Wang and
                  Doyen Sahoo and
                  Junnan Li and
                  Nan Jiang and
                  Tong Zhang and
                  Caiming Xiong and
                  Hanze Dong},
  title        = {A Minimalist Approach to {LLM} Reasoning: from Rejection Sampling
                  to Reinforce},
  journal      = {CoRR},
  volume       = {abs/2504.11343},
  year         = {2025},
  url          = {https://doi.org/10.48550/arXiv.2504.11343},
  doi          = {10.48550/ARXIV.2504.11343},
  eprinttype    = {arXiv},
  eprint       = {2504.11343},
  bibsource    = {dblp computer science bibliography, https://dblp.org}
}

@article{sample_filtering2,
  author       = {Kareem Amin and
                  Sara Babakniya and
                  Alex Bie and
                  Weiwei Kong and
                  Umar Syed and
                  Sergei Vassilvitskii},
  title        = {Escaping Collapse: The Strength of Weak Data for Large Language Model
                  Training},
  journal      = {CoRR},
  volume       = {abs/2502.08924},
  year         = {2025},
  url          = {https://doi.org/10.48550/arXiv.2502.08924},
  doi          = {10.48550/ARXIV.2502.08924},
  eprinttype    = {arXiv},
  eprint       = {2502.08924},
  bibsource    = {dblp computer science bibliography, https://dblp.org}
}

@inproceedings{ntk,
  author       = {Arthur Jacot and
                  Cl{\'{e}}ment Hongler and
                  Franck Gabriel},
  editor       = {Samy Bengio and
                  Hanna M. Wallach and
                  Hugo Larochelle and
                  Kristen Grauman and
                  Nicol{\`{o}} Cesa{-}Bianchi and
                  Roman Garnett},
  title        = {Neural Tangent Kernel: Convergence and Generalization in Neural Networks},
  booktitle    = {Advances in Neural Information Processing Systems 31: Annual Conference
                  on Neural Information Processing Systems 2018, NeurIPS 2018, December
                  3-8, 2018, Montr{\'{e}}al, Canada},
  pages        = {8580--8589},
  year         = {2018},
  url          = {https://proceedings.neurips.cc/paper/2018/hash/5a4be1fa34e62bb8a6ec6b91d2462f5a-Abstract.html},
  bibsource    = {dblp computer science bibliography, https://dblp.org}
}

@article{multi_temp,
  author       = {Haomin Zhuang and
                  Yujun Zhou and
                  Taicheng Guo and
                  Yue Huang and
                  Fangxu Liu and
                  Kai Song and
                  Xiangliang Zhang},
  title        = {Exploring Multi-Temperature Strategies for Token- and Rollout-Level
                  Control in {RLVR}},
  journal      = {CoRR},
  volume       = {abs/2510.08892},
  year         = {2025},
  url          = {https://doi.org/10.48550/arXiv.2510.08892},
  doi          = {10.48550/ARXIV.2510.08892},
  eprinttype    = {arXiv},
  eprint       = {2510.08892},
  bibsource    = {dblp computer science bibliography, https://dblp.org}
}

@article{open_o1,
  author       = {Aaron Jaech and
                  Adam Kalai and
                  Adam Lerer and
                  Adam Richardson and
                  Ahmed El{-}Kishky and
                  Aiden Low and
                  Alec Helyar and
                  Aleksander Madry and
                  Alex Beutel and
                  Alex Carney and
                  Alex Iftimie and
                  Alex Karpenko and
                  Alex Tachard Passos and
                  Alexander Neitz and
                  Alexander Prokofiev and
                  Alexander Wei and
                  Allison Tam and
                  Ally Bennett and
                  Ananya Kumar and
                  Andre Saraiva and
                  Andrea Vallone and
                  Andrew Duberstein and
                  Andrew Kondrich and
                  Andrey Mishchenko and
                  Andy Applebaum and
                  Angela Jiang and
                  Ashvin Nair and
                  Barret Zoph and
                  Behrooz Ghorbani and
                  Ben Rossen and
                  Benjamin Sokolowsky and
                  Boaz Barak and
                  Bob McGrew and
                  Borys Minaiev and
                  Botao Hao and
                  Bowen Baker and
                  Brandon Houghton and
                  Brandon McKinzie and
                  Brydon Eastman and
                  Camillo Lugaresi and
                  Cary Bassin and
                  Cary Hudson and
                  Chak Ming Li and
                  Charles de Bourcy and
                  Chelsea Voss and
                  Chen Shen and
                  Chong Zhang and
                  Chris Koch and
                  Chris Orsinger and
                  Christopher Hesse and
                  Claudia Fischer and
                  Clive Chan and
                  Dan Roberts and
                  Daniel Kappler and
                  Daniel Levy and
                  Daniel Selsam and
                  David Dohan and
                  David Farhi and
                  David Mely and
                  David Robinson and
                  Dimitris Tsipras and
                  Doug Li and
                  Dragos Oprica and
                  Eben Freeman and
                  Eddie Zhang and
                  Edmund Wong and
                  Elizabeth Proehl and
                  Enoch Cheung and
                  Eric Mitchell and
                  Eric Wallace and
                  Erik Ritter and
                  Evan Mays and
                  Fan Wang and
                  Felipe Petroski Such and
                  Filippo Raso and
                  Florencia Leoni and
                  Foivos Tsimpourlas and
                  Francis Song and
                  Fred von Lohmann and
                  Freddie Sulit and
                  Geoff Salmon and
                  Giambattista Parascandolo and
                  Gildas Chabot and
                  Grace Zhao and
                  Greg Brockman and
                  Guillaume Leclerc and
                  Hadi Salman and
                  Haiming Bao and
                  Hao Sheng and
                  Hart Andrin and
                  Hessam Bagherinezhad and
                  Hongyu Ren and
                  Hunter Lightman and
                  Hyung Won Chung and
                  Ian Kivlichan and
                  Ian O'Connell and
                  Ian Osband and
                  Ignasi Clavera Gilaberte and
                  Ilge Akkaya},
  title        = {OpenAI o1 System Card},
  journal      = {CoRR},
  volume       = {abs/2412.16720},
  year         = {2024},
  url          = {https://doi.org/10.48550/arXiv.2412.16720},
  doi          = {10.48550/ARXIV.2412.16720},
  eprinttype    = {arXiv},
  eprint       = {2412.16720},
  bibsource    = {dblp computer science bibliography, https://dblp.org}
}

@article{diffcult_filter,
  author       = {Sanghwan Bae and
                  Jiwoo Hong and
                  Min Young Lee and
                  Hanbyul Kim and
                  JeongYeon Nam and
                  Donghyun Kwak},
  title        = {Online Difficulty Filtering for Reasoning Oriented Reinforcement Learning},
  journal      = {CoRR},
  volume       = {abs/2504.03380},
  year         = {2025},
  url          = {https://doi.org/10.48550/arXiv.2504.03380},
  doi          = {10.48550/ARXIV.2504.03380},
  eprinttype    = {arXiv},
  eprint       = {2504.03380},
  bibsource    = {dblp computer science bibliography, https://dblp.org}
}
\bibliographystyle{icml2026}

\newpage
\appendix
\onecolumn
\crefalias{section}{appendix}

\section{Proof of \cref{thm:optimal_exact}}
\label{appendix:proof_optimal}

\begin{theorem}
The maximizer of \cref{eq:full_obj}, denoted as $\pi^*$, is given by
\begin{equation}
\pi^*(o|q) = \frac{1}{Z(q)}\, \pi_{\text{ref}}(o|q)\,
\exp\!\left( \frac{A(q,o)}{\beta} \right),
\end{equation}
where $Z(q)$ is the partition function ensuring normalization.
\end{theorem}
\begin{proof}
Because the expectation over $q$ is linear and the policy can depend arbitrarily on $q$, the maximization decomposes across queries.
For each fixed $q$, we omit it from notation for brevity and can independently maximize the per‑query objective as follows:
\[
J(\pi) = \int \pi(o) A(o) \, do - \beta \int \pi(o) \log \frac{\pi(o)}{\pi_{\text{ref}}(o)} \, do.
\]
We introduce a Lagrange multiplier $\lambda$ for the normalization constraint:
\[
\mathcal{L}[\pi, \lambda] = \int \pi(o) A(o) \, do - \beta \int \pi(o) \log \frac{\pi(o)}{\pi_{\text{ref}}(o)} \, do - \lambda \left( \int \pi(o) \, do - 1 \right).
\]

Taking the functional derivative with respect to $\pi(o)$ (pointwise) yields:
\[
\frac{\delta \mathcal{L}}{\delta \pi(o)} = A(o) - \beta \left( \log \frac{\pi(o)}{\pi_{\text{ref}}(o)} + 1 \right) - \lambda.
\]
Setting the derivative to zero for every $o$ gives:
\[
A(o) - \beta \left( \log \pi(o) - \log \pi_{\text{ref}}(o) + 1 \right) - \lambda = 0.
\]
We rearrange the above equation and obtain:
\[
\log \pi(o) = \log \pi_{\text{ref}}(o) + \frac{A(o)}{\beta} - \left( 1 + \frac{\lambda}{\beta} \right).
\]
We then exponentiate both sides:
\[
\pi(o) = \pi_{\text{ref}}(o) \, \exp\!\left( \frac{A(o)}{\beta} \right) \, \exp\!\left( -1 - \frac{\lambda}{\beta} \right).
\]

The factor $\exp(-1 - \lambda/\beta)$ is independent of $o$.
Due to $\int \pi(o) \, do = 1$, we have:
\[
\pi(o) = \frac{ \pi_{\text{ref}}(o) \, \exp\!\left( \frac{A(o)}{\beta} \right) }{ \int \pi_{\text{ref}}(o') \, \exp\!\left( \frac{A(o')}{\beta} \right) \, do' }.
\]
Restoring the explicit dependence on $q$, we obtain \cref{thm:optimal_exact}.
\end{proof}



\section{Proof of \cref{thm:batch_opt}}
\label{appendix_proof_batch_opt}

\begin{proof}
We begin by re-expressing the empirical objective $\hat{J}(\pi)$ at the mode level.
Given a query $q$, the objective is a combination of the sample-weighted advantage and the KL divergence relative to the reference policy:
$$
\hat{J}(\pi) = \sum_{i} \frac{N_i}{G} A(q, o_i) \log \pi(o_i|q) - \beta \sum_{i} \pi(o_i|q) \log \frac{\pi(o_i|q)}{\pi_{ref}(o_i|q)},
$$
where we treat $\pi$ as a functional optimization target over the discrete output space.

To find the maximizer $\hat{\pi}$, we introduce a Lagrange multiplier $\lambda$ for the normalization constraint $\sum_i \pi(o_i) = 1$.
The Lagrangian $\mathcal{L}$ is defined as:
$$
\mathcal{L}(\pi, \lambda) = \sum_{i} \left[ \frac{N_i A_i}{G} \log \pi(o_i) - \beta \pi(o_i) \log \frac{\pi(o_i)}{\pi_{ref}(o_i)} \right] - \lambda \left( \sum_{i} \pi(o_i) - 1 \right).
$$
Taking the functional derivative with respect to each $\pi(o_i)$ and setting it to zero yields:
$$
\frac{\partial \mathcal{L}}{\partial \pi(o_i)} = \frac{N_i A_i}{G \cdot \pi(o_i)} - \beta \left( \log \frac{\pi(o_i)}{\pi_{ref}(o_i)} + 1 \right) - \lambda = 0.
$$
Rearranging the terms to isolate the log-probability ratio:
$$
\log \frac{\pi(o_i)}{\pi_{ref}(o_i)} = \frac{N_i A_i}{\beta G \cdot \pi(o_i)} - \left( 1 + \frac{\lambda}{\beta} \right).
$$
Then, we have:
$$
\log \hat{\pi}(o_i) = \log \pi_{ref}(o_i) + \frac{N_i A_i}{\beta G} + C,
$$
where $C$ is a normalization constant independent of $o_i$.
Applying the constraint $\sum_i \hat{\pi}(o_i) = 1$ yields \cref{thm:batch_opt}.
\end{proof}

\section{Distributional Update Geometry}
\label{appendix:grad_interp}

\begin{theorem}
\label{thm:grad_interp}
Consider a single gradient ascent step $\theta^+ = \theta + \eta \nabla_\theta \hat{J}(\theta)$ on the empirical objective with a learning rate of $\eta$.
In the functional space of policies, let $\pi_t$ denote the policy at step t. 
Under a first-order Taylor expansion, the updated policy $\pi_{t+1}$ relates to the current policy $\pi_t$ and the batch-optimal policy $\hat{\pi}$ as:
\begin{equation}
\label{eq:interp_final}
\pi_{t+1}(o|q) \propto \pi_t(o|q)^{1 - \eta \beta} \cdot \hat{\pi}(o|q)^{\eta \beta}.
\end{equation}
\end{theorem}

\begin{proof}

The empirical objective $\hat{J}(\theta)$ is defined as:
$$\hat{J}(\theta) = \sum_{i} \frac{N_i}{G} A_i \log \pi_\theta(o_i) - \beta D_{KL}(\pi_\theta \| \pi_{\text{ref}}).$$
The batch-optimal policy $\hat{\pi}$ satisfies the stationary condition $\nabla_\pi \hat{J} = 0$, which yields $\hat{\pi}(o_i) = \frac{1}{Z'} \pi_{\text{ref}}(o_i) \exp(\frac{N_i A_i}{\beta G})$ according to \cref{thm:batch_opt}.
Taking the log:
$$\frac{N_i A_i}{\beta G} = \log \hat{\pi}(o_i) - \log \pi_{\text{ref}}(o_i) + \log Z'$$

The gradient of $\hat{J}$ with respect to the log-probability $w_i = \log \pi_\theta(o_i)$ is:
$$\frac{\partial \hat{J}}{\partial w_i} = \frac{N_i A_i}{G} - \beta (\log \pi_\theta(o_i) - \log \pi_{\text{ref}}(o_i) + 1).$$

Combining the above two equations:
$$\frac{\partial \hat{J}}{\partial w_i} = \beta \left( \log \hat{\pi}(o_i) - \log \pi_{\text{ref}}(o_i) + \log Z' \right) - \beta \log \pi_\theta(o_i) + \beta \log \pi_{\text{ref}}(o_i) - \beta = \beta \left( \log \hat{\pi}(o_i) - \log \pi_\theta(o_i) \right) + \text{constant}.
$$

In functional space, the update in log-space is:
$$\log \pi_{t+1}(o_i) = \log \pi_t(o_i) + \eta \frac{\partial \hat{J}}{\partial w_i} = \log \pi_t(o_i) + \eta \beta (\log \hat{\pi}(o_i) - \log \pi_t(o_i)) + C.$$

Rearranging the terms:
$$\log \pi_{t+1}(o_i) = (1 - \eta \beta) \log \pi_t(o_i) + \eta \beta \log \hat{\pi}(o_i) + C$$

Exponentiating both sides yields the geometric mean:
$$\pi_{t+1}(o_i) \propto \pi_t(o_i)^{1 - \eta \beta} \cdot \hat{\pi}(o_i)^{\eta \beta}.$$

\end{proof}

\section{Proof of \cref{thm:Z_bound}}
\label{appendix_proof_thm3}

\begin{proof}
We first derive the first bound.
By definition of $Z'(q)$ and partitioning modes into unsampled, correct, and incorrect sets, we have:
\begin{align*}
Z'(q)
&= \sum_{k \notin S^+ \cup S^-} \pi_{\mathrm{ref}}(o_k \mid q)
 + \sum_{i \in S^+} \pi_{\mathrm{ref}}(o_i \mid q)\,
   e^{\frac{N_i A_+}{\beta G}}
 + \sum_{j \in S^-} \pi_{\mathrm{ref}}(o_j \mid q)\,
   e^{\frac{N_j A_-}{\beta G}}.
\end{align*}
Using $e^x \ge 1 + x$, we obtain:
\begin{align*}
Z'(q)
&\ge \sum_{k \notin S^+ \cup S^-} \pi_{\mathrm{ref}}(o_k \mid q)
 + \sum_{i \in S^+} \pi_{\mathrm{ref}}(o_i \mid q)
   \Big(1 + \frac{N_i A_+}{\beta G}\Big)
 + \sum_{j \in S^-} \pi_{\mathrm{ref}}(o_j \mid q)
   \Big(1 + \frac{N_j A_-}{\beta G}\Big) \\
&= \sum_{\text{all } k} \pi_{\mathrm{ref}}(o_k \mid q)
 + \frac{1}{\beta G}
   \Big[
    A_+ \sum_{i \in S^+} N_i \,\pi_{\mathrm{ref}}(o_i \mid q)
  + A_- \sum_{j \in S^-} N_j \,\pi_{\mathrm{ref}}(o_j \mid q)
   \Big] \\
&= 1 + \frac{1}{\beta G}
\Big[
 A_+ \sum_{i \in S^+} N_i \,\pi_{\mathrm{ref}}(o_i \mid q)
 - |A_-| \sum_{j \in S^-} N_j \,\pi_{\mathrm{ref}}(o_j \mid q)
\Big],
\end{align*}
which proves the first bound.

We now specialize to normalized binary advantages.
Substitute $A_+ = (1 - p^+)/\sigma$ and $A_- = -p^+/\sigma$ into the above bound:
\begin{align*}
Z'(q)
&\ge 1 + \frac{1}{\beta G \sigma}
\Big[
 (1 - p^+) \sum_{i \in S^+} N_i \,\pi_{\mathrm{ref}}(o_i \mid q)
 - p^+ \sum_{j \in S^-} N_j \,\pi_{\mathrm{ref}}(o_j \mid q)
\Big].
\end{align*}
Using $\sum_{i \in S^+} N_i = G p^+$ and $\sum_{j \in S^-} N_j = G (1 - p^+)$, we can express the weighted averages:
\[
\bar{\pi}^+ = \frac{\sum_{i \in S^+} N_i \,\pi_{\mathrm{ref}}(o_i \mid q)}{\sum_{i \in S^+} N_i},
\qquad
\bar{\pi}^- = \frac{\sum_{j \in S^-} N_j \,\pi_{\mathrm{ref}}(o_j \mid q)}{\sum_{j \in S^-} N_j}.
\]
Then we have:
\begin{align*}
Z'(q)
&\ge 1 + \frac{1}{\beta \sigma}
\Big[
 (1 - p^+) p^+ \,\bar{\pi}^+
 - p^+ (1 - p^+) \,\bar{\pi}^-
\Big] \\
&= 1 + \frac{p^+ (1 - p^+)}{\beta \sigma}
\big(\bar{\pi}^+ - \bar{\pi}^-\big).
\end{align*}
By definition,
\[
\pi_{\min}^+ \le \bar{\pi}^+,
\qquad
\bar{\pi}^- \le \pi_{\max}^-,
\]
so
\[
\bar{\pi}^+ - \bar{\pi}^- \;\ge\; \pi_{\min}^+ - \pi_{\max}^- \;=\; \Delta_{\pi}.
\]
Therefore, we see:
\[
Z'(q)
\;\ge\;
1 + \frac{p^+ (1 - p^+)}{\beta \sigma}\,\Delta_{\pi}.
\]
If $\Delta_{\pi} > 0$, the right‑hand side is greater than 1.
For any unsampled mode $o_k$ with $N_k = 0$, \cref{thm:batch_opt} gives
\[
\hat{\pi}(o_k \mid q)
= \frac{\pi_{\mathrm{ref}}(o_k \mid q)}{Z'(q)} < \pi_{\mathrm{ref}}(o_k \mid q).
\]
Thus, all unsampled modes are uniformly suppressed.
\end{proof}

\section{Background for \cref{sec2_3}}
\label{appendix_background}

The derivation presented here is largely adapted from the Neural Tangent Kernel framework.
We now provide the derivation for the parameter update form $\Delta \theta$ and the induced logit shift $\Delta f$ used in the main text.
This derivation relies on the local properties of the neural network function under small parameter updates.
For a small optimization step, the change in the model parameters $\Delta \theta$ is small.
Consequently, the change in the network output can be approximated by the first-order Taylor expansion around the current parameters $\theta$:
$$
f(x; \theta + \Delta \theta) \approx f(x; \theta) + \nabla_\theta f(x; \theta)^\top \Delta \theta.
$$
Consider a training batch with target logit shifts given by the vector $\mathbf{y}$.
To achieve this shift locally, the parameter update $\Delta \theta$ must satisfy the linearized constraint:
$$
\mathbf{J}_q \Delta \theta = \mathbf{y},
$$
where $\mathbf{J}_q$ is the Jacobian matrix of the batch samples.
Since the number of parameters is typically much larger than the batch size, this system is under-determined.
Standard gradient-based optimizers naturally update the parameters within the subspace spanned by the data gradients (the row space of $\mathbf{J}_q$).
This corresponds to finding the solution with the minimum Euclidean norm $\|\Delta \theta\|_2$, representing the steepest direction to fit the local target:
$$
\Delta \theta = \operatorname*{arg\,min}_{\delta} \|\delta\|_2^2 \quad \text{s.t.} \quad \mathbf{J}_q \delta = \mathbf{y}.
$$
The closed-form solution to this problem is given by the Moore-Penrose pseudoinverse:
$$
\Delta \theta = \mathbf{J}_q^\top (\mathbf{J}_q \mathbf{J}_q^\top)^{-1} \mathbf{y}.
$$
Here, the matrix $\mathbf{K}_{qq} = \mathbf{J}_q \mathbf{J}_q^\top$ captures the local alignment between gradients of the batch samples.
Once $\Delta \theta$ is determined, the change in the logit value for any other query-output pair $(q', o')$ is governed by the same local linearization:
$$
\Delta f(q', o') \approx \nabla_\theta f(q', o')^\top \Delta \theta.
$$
Substituting the expression for $\Delta \theta$, we obtain:
$$
\Delta f(q', o') = \nabla_\theta f(q', o')^\top \mathbf{J}_q^\top (\mathbf{J}_q \mathbf{J}_q^\top)^{-1} \mathbf{y} = \mathbf{k}_{q'}^\top \mathbf{K}_{qq}^{-1} \mathbf{y},
$$
where $\mathbf{k}_{q'} = \mathbf{J}_q \nabla_\theta f(q', o')$ represents the vector of gradient dot products (kernel values) between the new query and the training batch.
This confirms that within a small update step, the propagation of probability mass is dictated by gradient similarity.

\section{Assumption Validation}
\label{appendix_ass_valid}

We empirically validate Assumptions~\ref{assump:nonneg_grad} and~\ref{assump:diagonal_dominance} on Qwen3‑0.6B.
We first randomply extract 1000 pairs of questions $(q, q')$ from GSM8K and, for each query, sample 8 responses.
For $(q, o)$ and $(q', o')$, we compute the inner product of their logit gradients $\nabla_\theta f(q, o)^\top \nabla_\theta f(q', o')$.
Across all such pairs, the mean inner product is 1467.50 with a standard deviation of 363.97, indicating that gradients for semantically related outputs are strongly and consistently positively aligned, in line with Assumption~\ref{assump:nonneg_grad}.

\begin{figure}
    \centering
    \includegraphics[width=0.5\linewidth]{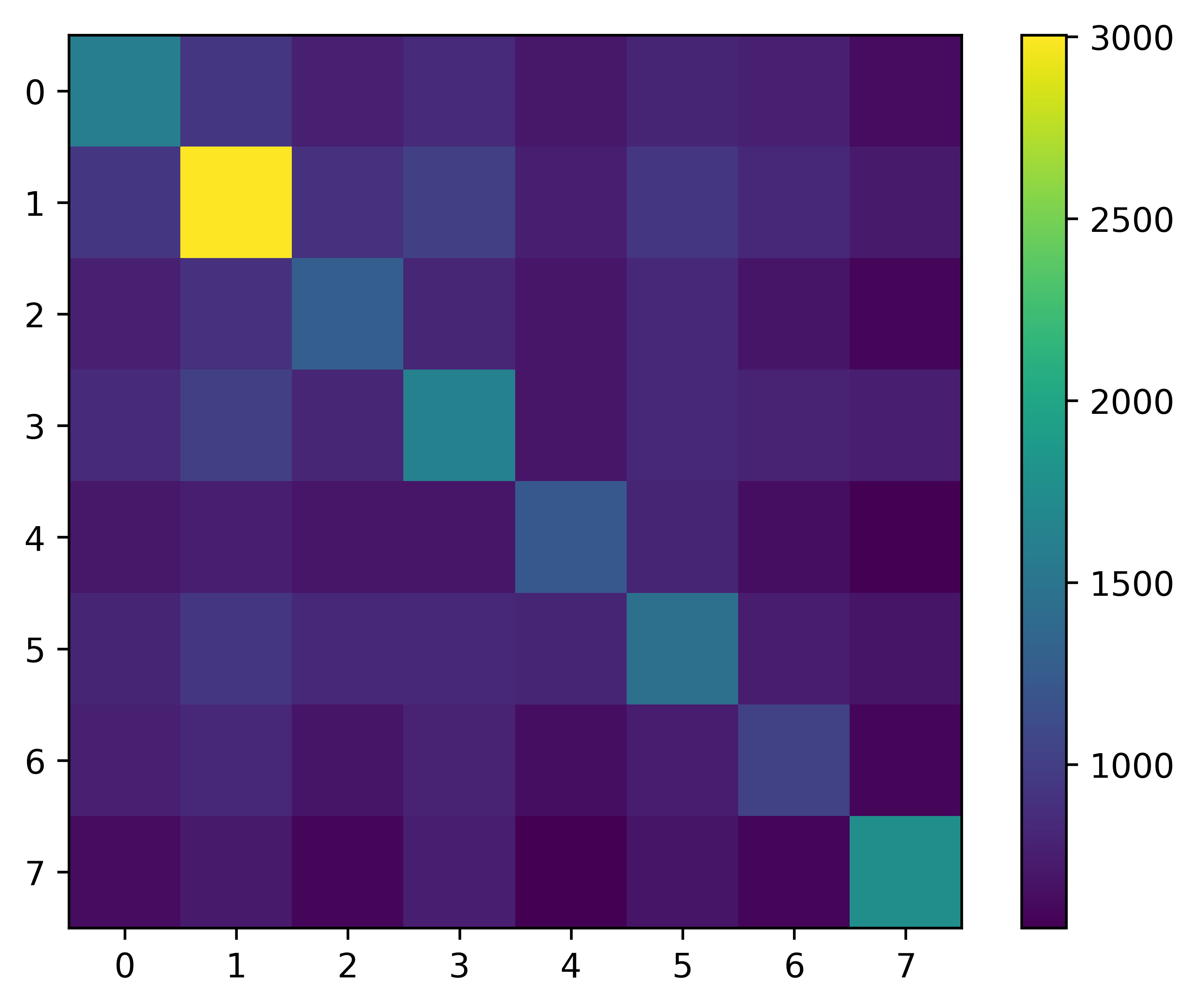}
    \caption{
    Empirical validation of diagonal dominance on Qwen3‑0.6B using GSM8K.
    Empirical kernel matrix $\mathbf{K}_{qq}$, where diagonal entries (self‑alignment) are larger than off‑diagonal entries (cross‑alignment).
    }
    \label{fig:qwen06B_kernel}
\end{figure}

We then estimate the empirical kernel matrix $\mathbf{K}_{qq} = \mathbf{J}_q \mathbf{J}_q^\top$ for $q$.
As shown in \cref{fig:qwen06B_kernel}, the diagonal entries (self‑alignment) are substantially larger than the off‑diagonal entries (cross‑alignment), corroborating the diagonal‑dominance condition in Assumption~\ref{assump:diagonal_dominance}.

\section{Proof of \cref{thm:logit_bounds}}
\label{appendix_proof_thm4}

\begin{theorem}
Let $\mathcal{A}_{\text{sum}} \triangleq \sum_{s=1}^G y_s = \frac{1}{\beta G} \sum_{s=1}^G N_s |A(q, o^{(s)})|.$
The induced logit shift $\Delta f(q', o')$ satisfies the following bounds based on whether $o'$ appears in the training batch $\{o^{(i)}, i=1,\cdots,G\}$:
\begin{itemize}[leftmargin=*,topsep=1pt]
    \item  If $o' \notin \{o^{(s)}\}_{s=1}^G$, the logit increase is upper-bounded by:
    $$
    \Delta f(q', o') \leq  \frac{\eta \rho_{\max} \mathcal{A}_{\text{sum}}}{\lambda_{\min} + (G-1)\rho_{\min}}.
    $$
    \item If $o' = o^{(k)}$ for some $k \in \{1, \dots, G\}$, the logit increase is lower-bounded by
    $$
    \Delta f(q', o') = \Delta f(q', o^{(k)}) \geq \frac{\eta N_{o'} (\lambda_{\min} - \rho_{\min})}{\lambda_{\max} - \rho_{\max}} \left( y_k - \frac{\rho_{\max} \mathcal{A}_{\text{sum}}}{\lambda_{\max} + (G-1)\rho_{\max}} \right).
    $$
\end{itemize}
\end{theorem}

\begin{proof}

We define the dual coefficient vector $\boldsymbol{\alpha} = \mathbf{K}_{qq}^{-1} \mathbf{y}$.
We proceed by first bounding the entries of $\boldsymbol{\alpha}$.

\textbf{Step 1: Bounds on $\boldsymbol{\alpha}$.}
Let $\mathbf{M}(\lambda, \rho)= (\lambda - \rho)\mathbf{I} + \rho \mathbf{1}\mathbf{1}^\top,$ where diagonal entries are $\lambda > 0$ and off-diagonal entries are $\rho > 0$.
Consider $\boldsymbol{\alpha}(\lambda, \rho) = \mathbf{M}(\lambda, \rho)^{-1} \mathbf{y}$.
Applying the Sherman-Morrison formula\footnote{$(\mathbf{A} + \mathbf{u}\mathbf{v}^\top)^{-1} = \mathbf{A}^{-1} - \frac{\mathbf{A}^{-1}\mathbf{u}\mathbf{v}^\top \mathbf{A}^{-1}}{1 + \mathbf{v}^\top \mathbf{A}^{-1} \mathbf{u}}$.}, the inverse $\mathbf{M}(\lambda, \rho)^{-1}$ applied to $\mathbf{y}$ yields:
$$
\boldsymbol{\alpha}(\lambda, \rho) = \mathbf{M}(\lambda, \rho)^{-1} \mathbf{y} = \frac{1}{\lambda - \rho} \left( \mathbf{I} - \frac{\rho}{\lambda + (G-1)\rho} \mathbf{1}\mathbf{1}^\top \right) \mathbf{y}.
$$
The $k$-th coefficient of $\boldsymbol{\alpha}(\lambda, \rho)$ is given by:
\begin{align}
\alpha_k(\lambda, \rho) = \frac{1}{\lambda - \rho} \left( y_k - \frac{\rho \mathcal{A}_{\text{sum}}}{\lambda + (G-1)\rho} \right). \nonumber
\end{align}
Summing over all $k$, the interaction terms cancel out, simplifying to a dependence on the total mass:
$$
\sum_{s} \alpha_s(\lambda, \rho) = \mathbf{1}^\top \mathbf{M}(\lambda, \rho)^{-1} \mathbf{y} = \frac{\mathcal{A}_{\text{sum}}}{\lambda + (G-1)\rho}.
$$
Utilizing non-negativity of $y_i$, the spectral lower bound $\mathbf{K}_{qq} \succeq \mathbf{M}(\lambda_{\min}, \rho_{\min}) \succ 0$ and the order-reversing property of the matrix inverse, the sum of the coefficients is upper-bounded:
$$
\sum_{s=1}^G \alpha_s = \mathbf{1}^\top \mathbf{K}_{qq}^{-1} \mathbf{y} \leq \mathbf{1}^\top \mathbf{M}(\lambda_{\min}, \rho_{\min})^{-1} \mathbf{y}  = \frac{ \mathcal{A}_{\text{sum}}}{\lambda_{\min} + (G-1)\rho_{\min}}.
$$
Similarly, we have:
$$
\sum_{s=1}^G \alpha_s \geq \mathbf{1}^\top \mathbf{M}(\lambda_{\max}, \rho_{\max})^{-1} \mathbf{y} = \frac{\mathcal{A}_{\text{sum}}}{\lambda_{\max} + (G-1)\rho_{\max}}.
$$
For individual dual coefficients $\alpha_k$, we have the following two-sided bounds:
$$
\frac{1}{\lambda_{\max} - \rho_{\max}} \left( y_k - \frac{\rho_{\max} \mathcal{A}_{\text{sum}}}{\lambda_{\max} + (G-1)\rho_{\max}} \right) \leq \alpha_k \leq \frac{1}{\lambda_{\min} - \rho_{\min}} \left( y_k - \frac{\rho_{\min} \mathcal{A}_{\text{sum}}}{\lambda_{\min} + (G-1)\rho_{\min}} \right).
$$

\textbf{Step 2: Bounds on Logit Shift $\Delta f(q', o')$.} 
The logit change is given by:
$$
\Delta f(q', o') = \sum_{s} \alpha_s k((q', o'), (q, o^{(s)})).
$$
According to \cref{assump:tranfer_kernel}, we have $[\mathbf{k}_{q'}]_s =  k((q', o'), (q, o^{(s)})) \approx \eta K((q, o'), (q, o^{(s)}))$.

\textbf{Case 1: $o' \notin \{o^{(s)}\}_{s=1}^G$.}
Intuitively, for any distinct modes $o_a \neq o_b$, their kernel similarity is bounded by $\rho_{\max}$.
Intuitively, the batch $\{o^{(s)}\}$ captures the densest region of the optimization landscape, i.e., the high-probability modes that share the most structural features and thus exhibit the highest gradient overlap.
Consequently, any unseen mode $o' \notin \{o^{(s)}\}$ resides in the distribution's tail and is expected to have a lower gradient correlation with the batch samples than the samples have with each other, ensuring the bound holds.
Then, we have:
$$
[\mathbf{k}_{q'}]_s = k((q', o'), (q, o^{(s)})) \approx \eta K((q, o'), (q, o^{(s)})) \leq \eta \rho_{\max}.
$$
Using bounds derived in Step 1, we obtain:
$$
\Delta f(q', o') \leq \sum_{s=1}^G \alpha_s (\eta \rho_{\max}) = \eta \rho_{\max} \sum_{s=1}^G \alpha_s \leq \frac{\eta \rho_{\max} \mathcal{A}_{\text{sum}}}{\lambda_{\min} + (G-1)\rho_{\min}}.
$$

\textbf{Case 2: $o' \in \{o^{(s)}\}_{s=1}^G$.}
Let $S_{o'} = \{ s \mid o^{(s)} = o' \}$ be the set of indices where the sample $o'$ appears in the batch, with cardinality $|S_{o'}| = N_{o'}$.
The logit shift accumulates contributions from all $N_{o'}$ occurrences (self-terms) and the remaining $G - N_{o'}$ samples (cross-terms):
$$
\Delta f(q', o') = \sum_{s \in S_{o'}} \alpha_s [\mathbf{k}_{q'}]_s + \sum_{s \notin S_{o'}} \alpha_s [\mathbf{k}_{q'}]_s.
$$

To derive a lower bound, we assume the stiffest kernel parameters ($\lambda_{\max}, \rho_{\max}$) for $\boldsymbol{\alpha}$ (minimizing gradient magnitude) and the loosest parameters ($\lambda_{\min}, \rho_{\min}$) for transfer.
Substituting $[\mathbf{k}_{q'}]_k \ge \eta \lambda_{\min}$ and $[\mathbf{k}_{q'}]_j \ge \eta \rho_{\min}$:
\begin{align}
\Delta f(q', o') &\geq \eta \lambda_{\min} \sum_{s \in S_{o'}} \alpha_s + \eta \rho_{\min} \sum_{s \notin S_{o'}} \alpha_s \nonumber \\ 
&= \eta \lambda_{\min} \sum_{s \in S_{o'}} \alpha_s + \eta \rho_{\min} \left( \sum_{all} \alpha_s - \sum_{s \in S_{o'}} \alpha_s \right) \nonumber \\ 
&= \eta (\lambda_{\min} - \rho_{\min}) \sum_{s \in S_{o'}} \alpha_s + \eta \rho_{\min} \mathcal{A}_{\text{sum}}.
\nonumber \end{align}

Next, we substitute the exact form of $\alpha_k$ derived in Step 1, using the stiff parameters $\lambda_{\max}$ and $\rho_{\max}$:
$$
\alpha_k \geq \frac{1}{\lambda_{\max} - \rho_{\max}} \left( y_k - \frac{\rho_{\max} \mathcal{A}_{\text{sum}}}{\lambda_{\max} + (G-1)\rho_{\max}} \right).
$$

Summing this over the $N_{o'}$ indices in $S_{o'}$:
\begin{align}
\sum_{s \in S_{o'}} \alpha_s &\geq \frac{1}{\lambda_{\max} - \rho_{\max}} \left( \sum_{s \in S_{o'}} y_s - N_{o'} \frac{\rho_{\max} \mathcal{A}_{\text{sum}}}{\lambda_{\max} + (G-1)\rho_{\max}} \right). \nonumber
\end{align}

Substituting this back into the logit shift equation and dropping the non-negative term $\eta \rho_{\min} \mathcal{A}_{\text{sum}}$:
$$
\Delta f(q', o') \geq  \frac{\eta N_{o'} (\lambda_{\min} - \rho_{\min})}{\lambda_{\max} - \rho_{\max}} \left( y_k - \frac{\rho_{\max} \mathcal{A}_{\text{sum}}}{\lambda_{\max} + (G-1)\rho_{\max}} \right).
$$

\end{proof}

\section{Proof of \cref{corollary:ratio}}
\label{appendix_proof_thm5}

\begin{corollary}
Define the ratio $\mathcal{R}$ of the lower bound in Case 2 to the upper bound in Case 1:
$$
\mathcal{R} \triangleq \frac{\lambda_{\min} - \rho_{\min}}{\lambda_{\max} - \rho_{\max}} \cdot \left[ \frac{N_{o'} y_k}{\mathcal{A}_{\text{sum}}} \cdot \left( \frac{\lambda_{\min} + (G-1)\rho_{\min}}{\rho_{\max}} \right) - \frac{\lambda_{\min} + (G-1)\rho_{\min}}{\lambda_{\max} + (G-1)\rho_{\max}} \right].
$$
Under the assumption of a well-conditioned kernel (where $\lambda_{\min} \approx \lambda_{\max} \approx \lambda$ and $\rho_{\min} \approx \rho_{\max} \approx \rho$), this ratio simplifies to:
$$\mathcal{R} \approx N_{o'} \left[ \frac{y_k}{\mathcal{A}_{\text{sum}}} \left( \frac{\lambda}{\rho} + G - 1 \right) - 1 \right].
$$
\end{corollary}

\begin{proof}
Dividing the lower bound of Case 2 by the upper bound of Case 1 in \cref{thm:logit_bounds}:
\begin{align}
\mathcal{R} &= \frac{ \frac{\eta N_{o'} (\lambda_{\min} - \rho_{\min})}{\lambda_{\max} - \rho_{\max}} \left( y_k - \frac{\rho_{\max} \mathcal{A}_{\text{sum}}}{\lambda_{\max} + (G-1)\rho_{\max}} \right) }{ \frac{\eta \rho_{\max} \mathcal{A}_{\text{sum}}}{\lambda_{\min} + (G-1)\rho_{\min}} } \nonumber \\
&= \frac{\lambda_{\min} - \rho_{\min}}{\lambda_{\max} - \rho_{\max}} \cdot \left[ \frac{N_{o'} y_k}{\mathcal{A}_{\text{sum}}} \cdot \left( \frac{\lambda_{\min} + (G-1)\rho_{\min}}{\rho_{\max}} \right) - \frac{\lambda_{\min} + (G-1)\rho_{\min}}{\lambda_{\max} + (G-1)\rho_{\max}} \right]. \nonumber
\end{align}

Assuming uniform parameters $\lambda$ and $\rho$, the stiffness damping factor $\frac{\lambda-\rho}{\lambda-\rho}$ becomes 1.
The common term $\eta$ cancels out.
We are left with:
\begin{align}
\mathcal{R} \approx \frac{ N_{o'} \left( y_k - \frac{\rho \mathcal{A}_{\text{sum}}}{\lambda + (G-1)\rho} \right) }{ \frac{\rho \mathcal{A}_{\text{sum}}}{\lambda + (G-1)\rho} } = N_{o'} \left[ \frac{y_k}{\mathcal{A}_{\text{sum}}} \left( \frac{\lambda}{\rho} + G - 1 \right) - 1 \right]. \nonumber
\end{align}

\end{proof}

\section{Supplementary Experiment for \cref{sec2_4}}
\label{appendix_sup_exp_sec24}

\begin{figure*}[!t]
    \centering
    \begin{subfigure}[b]{0.45\textwidth}
        \centering
        \includegraphics[width=\textwidth]{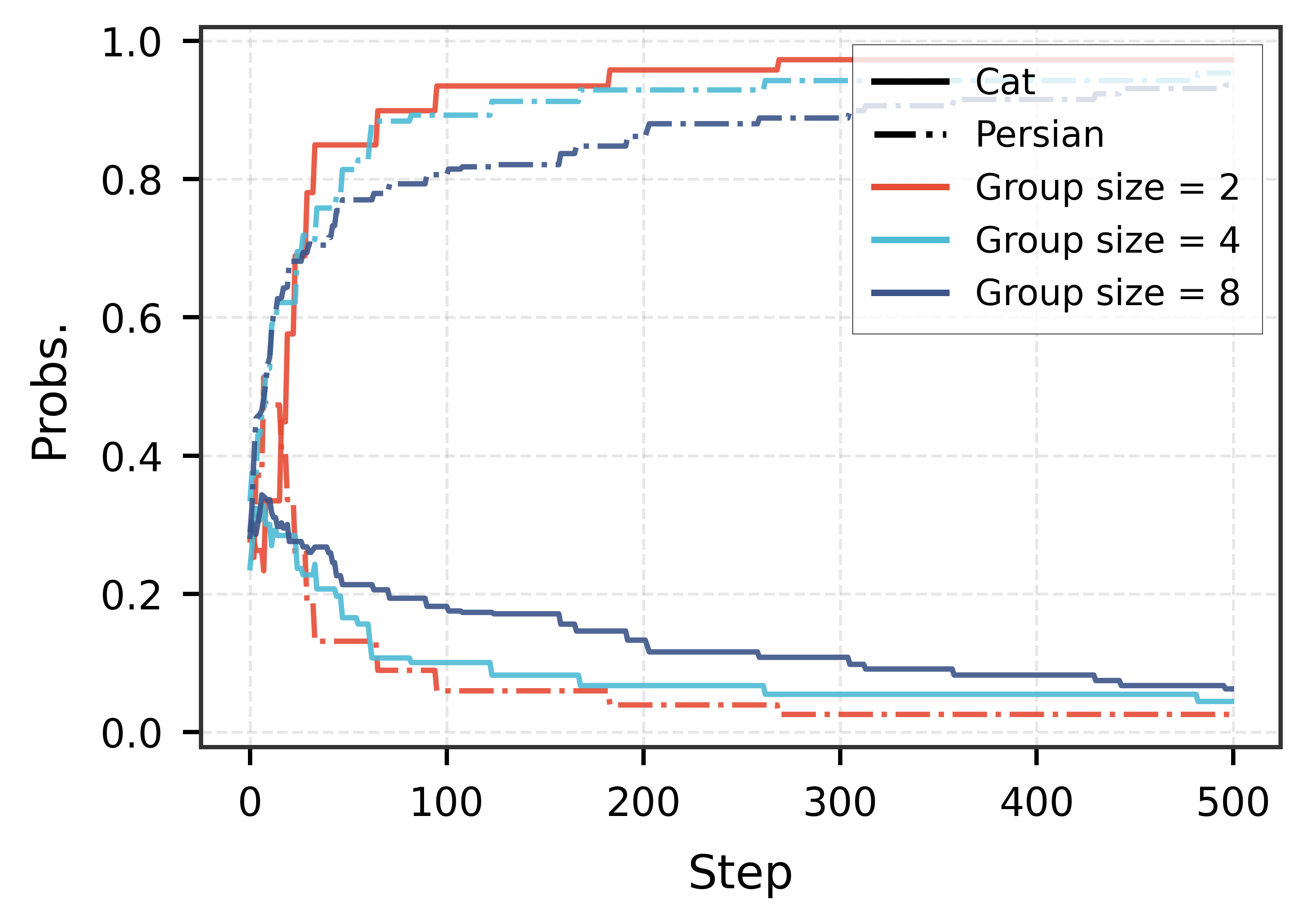}
        \caption{RLOO}
    \end{subfigure}
    \hfill
    \begin{subfigure}[b]{0.45\textwidth}
        \centering
        \includegraphics[width=\textwidth]{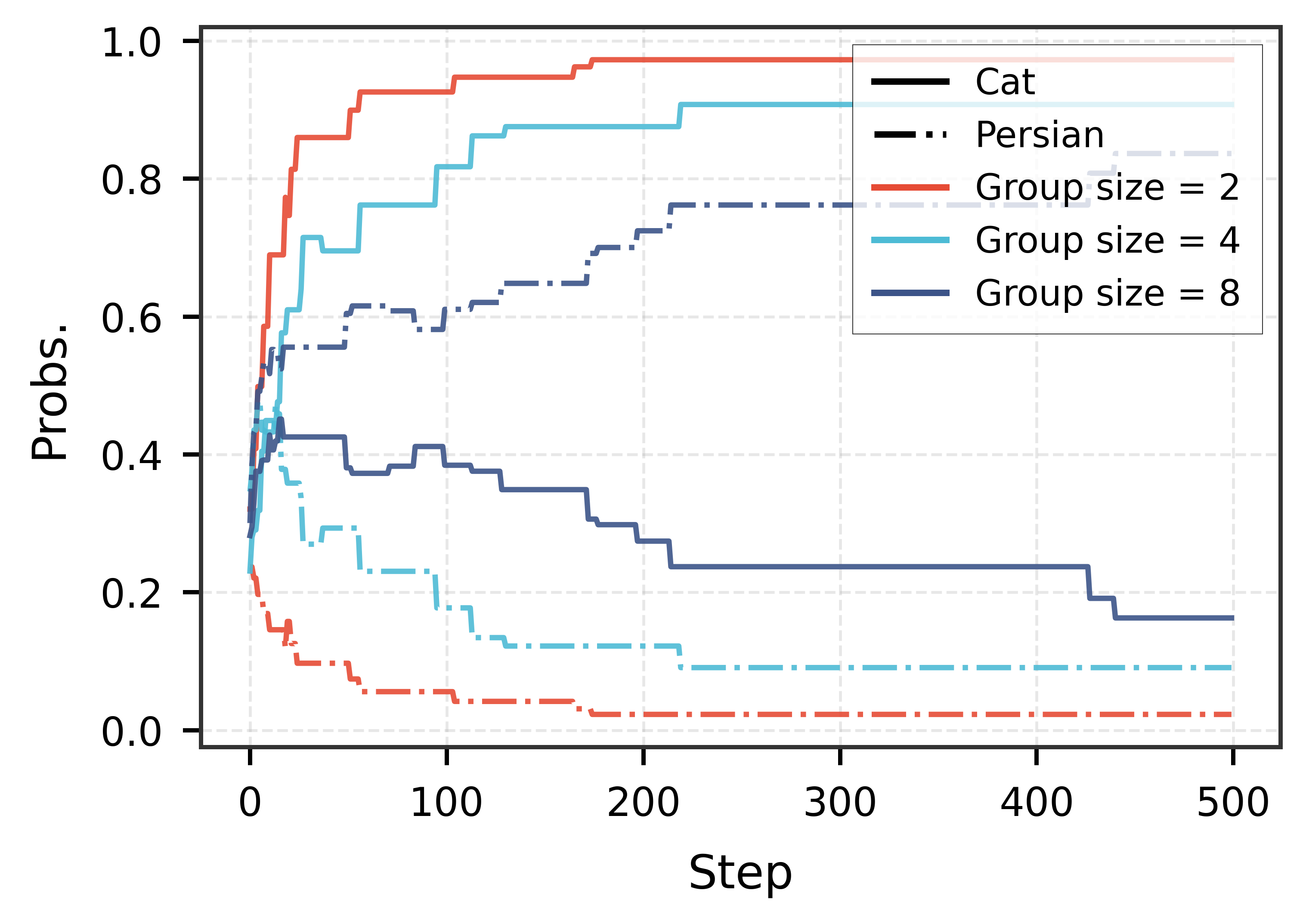}
        \caption{Reinforce++ with baseline}
    \end{subfigure}
    \caption{The effectiveness of RLOO and Reinforce++ with baseline on over-sharpening.}
    \label{fig:adv_estimators_ablation}
\end{figure*}

\textbf{Advanced Advantage Estimators.}
To ensure that the phenomenon of over-sharpening is not an artifact of simplistic advantage estimation, we extend our analysis to include RLOO and Reinforce++ with baseline.
As illustrated in \cref{fig:adv_estimators_ablation}, while these methods can slightly delay the onset of collapse compared to raw rewards, the policy eventually collapses onto a single mode.

\begin{figure*}[!t]
    \centering
    \begin{subfigure}[b]{0.45\textwidth}
        \centering
        \includegraphics[width=\textwidth]{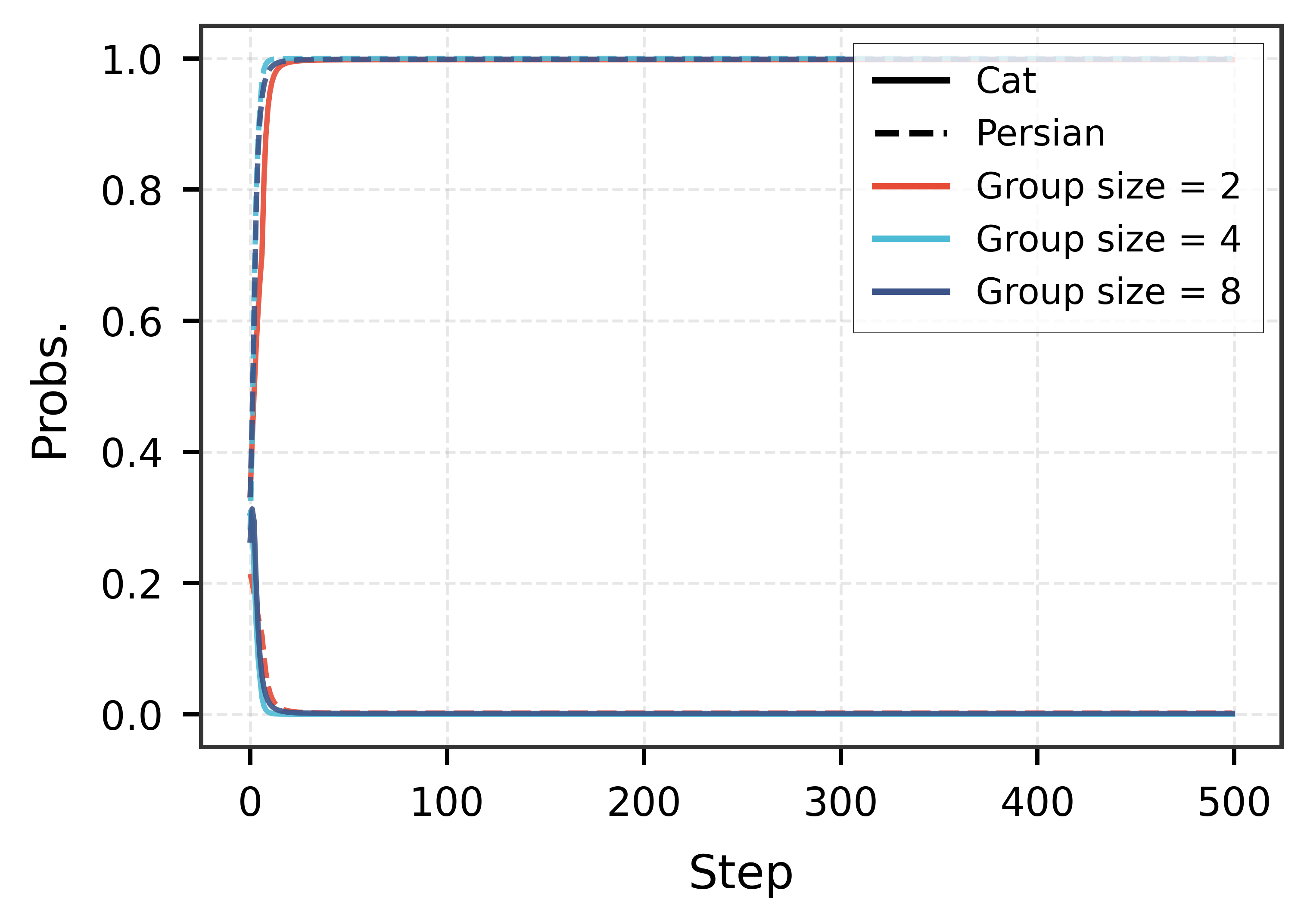}
        \caption{MI}
    \end{subfigure}
    \hfill
    \begin{subfigure}[b]{0.45\textwidth}
        \centering
        \includegraphics[width=\textwidth]{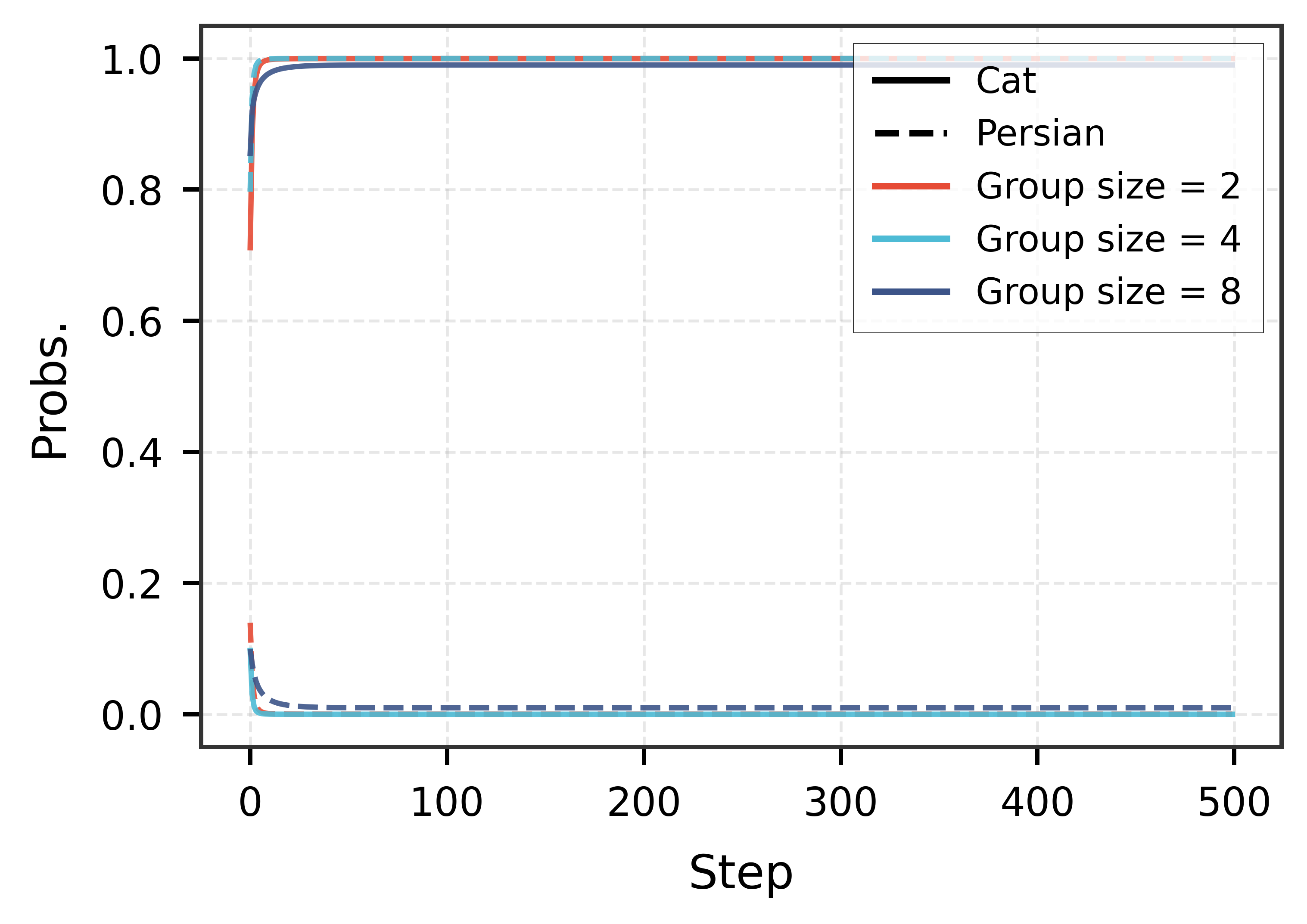}
        \caption{AdamW}
    \end{subfigure}
    \caption{Effect of different optimizers on over-sharpening.}
    \label{fig:optimizer_impact}
\end{figure*}

\textbf{Optimizer effect.}
As shown in \cref{fig:optimizer_impact}, both Momentum optimizer and AdamW optimizer significantly accelerate the rate of collapse compared to standard SGD.
We see that, for Momentum optimizer, the policy collapses almost instantaneously (within 20 steps) regardless of group size.
This occurs because the momentum buffer accumulates gradients from early samples.
Once sampling bias creates a slight preference for one mode (e.g., Cat), the history buffer sustains the update direction even if subsequent batches are more balanced.
For AdamW optimizer, we observe a worse trend.
By normalizing updates based on the second moment of the gradients, AdamW amplifies the effective learning rate, allowing the model to maximize the likelihood of the initially sampled modes more aggressively.

\section{Supplementary Material for \cref{sec3}}
\label{appendix_sup_exp_sec3}

\cref{alg:rlvr_calibrated} summarizes the complete pipeline of our methods.

\begin{algorithm}[t]
\caption{Inverse‑success Advantage Calibration + Distribution-Level Calibration}
\label{alg:rlvr_calibrated}
\small
\begin{algorithmic}[1]
\Require policy $\pi_\theta$, reference $\pi_{\text{ref}}$, optional memory $m_\phi$, batch size $B$, group size $G$, calibration strength $\alpha$, memory weight $\mu$
\While{not converged}
    \State Sample a mini-batch of queries $\{q_b\}_{b=1}^B$ from $\mathcal{D}$;
    \For{each query $q_b$}
        \If{memory $m_\phi$ is used}
            \State Compute calibrated logits $\tilde{f}(q_b, \cdot) = f_\theta(q_b, \cdot) - \mu f_\phi(q_b, \cdot)$;
            \State Sample $G$ trajectories $\{o_{b,s}\}_{s=1}^G \sim \text{softmax}(\tilde{f}(q_b,\cdot))$;
        \Else
            \State Sample $G$ trajectories $\{o_{b,s}\}_{s=1}^G \sim \pi_\theta(\cdot\mid q_b)$;
        \EndIf
        \State Compute verifiable rewards $r_{b,s}$ and advantages $A_{b,s}$;
        \State Identify positive indices $S_b^+ = \{s : A_{b,s} > 0\}$, negatives $S_b^-$;
        \State Compute success count $|S_b^+|$;
        \For{$s \in S_b^+$} \Comment{inverse-success calibration}
            \State $\tilde{A}_{b,s} \gets A_{b,s} \cdot (G - |S_b^+|)^\alpha$;
        \EndFor
        \For{$s \in S_b^-$} \Comment{leave negatives unscaled}
            \State $\tilde{A}_{b,s} \gets A_{b,s}$; 
        \EndFor
    \EndFor
    \State Update $\theta$ using calibrated advantages $\tilde{A}_{b,s}$ and KL regularization to $\pi_{\text{ref}}$;
    \If{memory $m_\phi$ is used}
        \State Update $m_\phi$ on sampled $(q_b,o_{b,s})$ pairs to track empirical frequencies;
    \EndIf
\EndWhile
\end{algorithmic}
\end{algorithm}

\begin{figure*}[!t]
    \centering
    \begin{subfigure}[b]{0.45\textwidth}
        \centering
        \includegraphics[width=\textwidth]{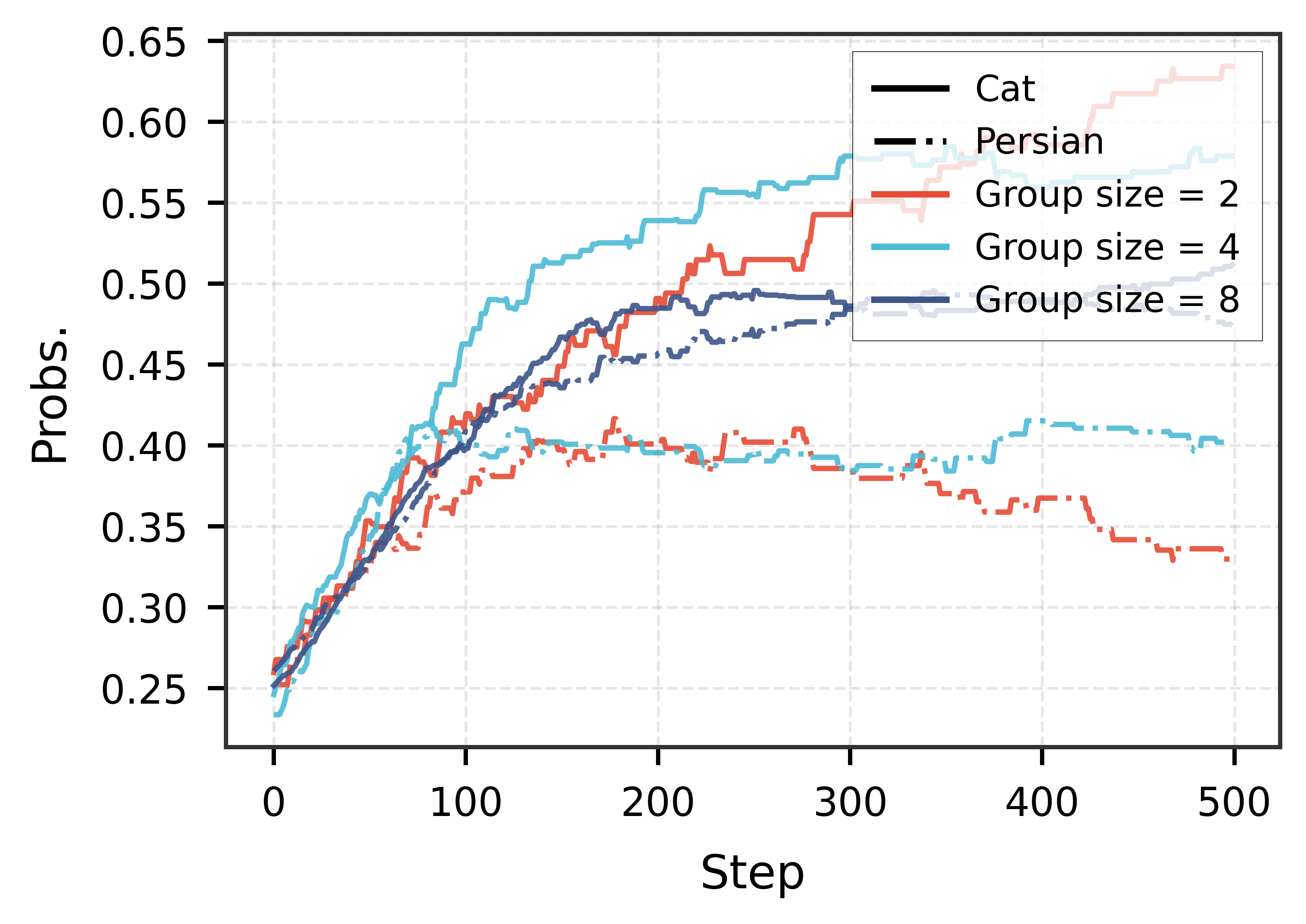}
        \caption{Inverse-success Calibration in SGD}
    \end{subfigure}
    \hfill
    \begin{subfigure}[b]{0.45\textwidth}
        \centering
        \includegraphics[width=\textwidth]{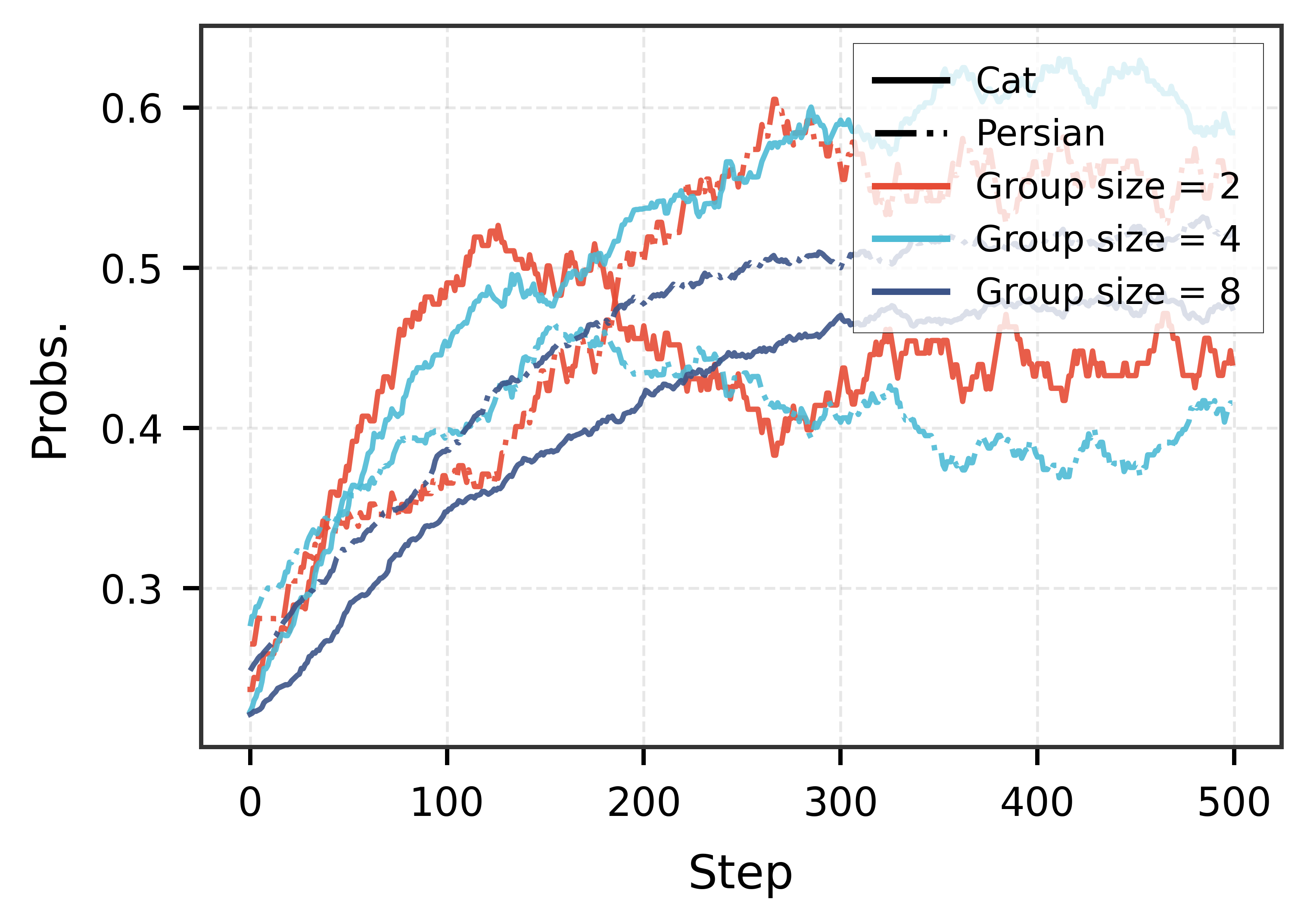}
        \caption{Inverse-success Calibration + Dist. Calib. in SGD}
    \end{subfigure} \\
    \begin{subfigure}[b]{0.45\textwidth}
        \centering
        \includegraphics[width=\textwidth]{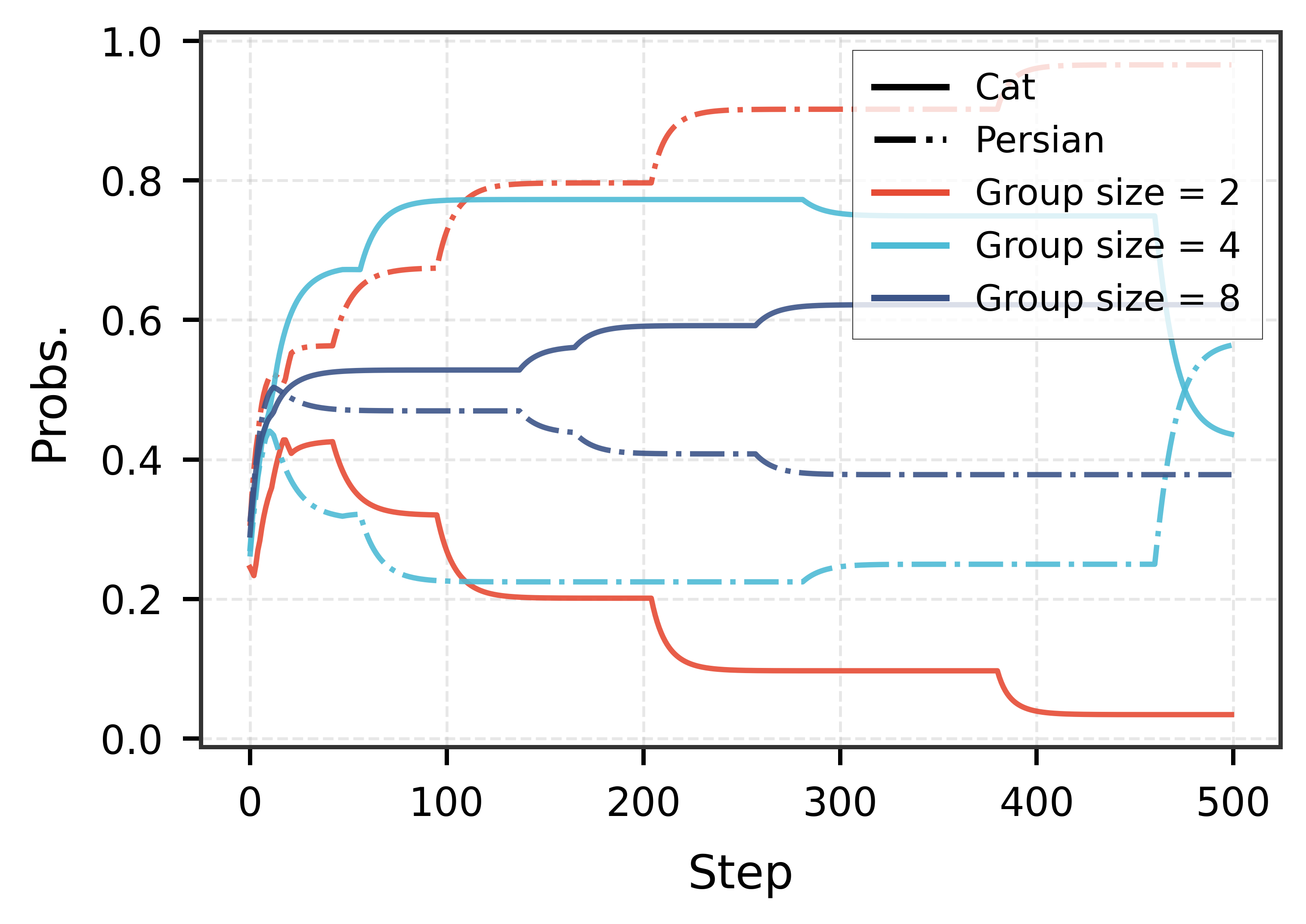}
        \caption{Inverse-success Calibration in AdamW}
    \end{subfigure}
    \hfill
    \begin{subfigure}[b]{0.45\textwidth}
        \centering
        \includegraphics[width=\textwidth]{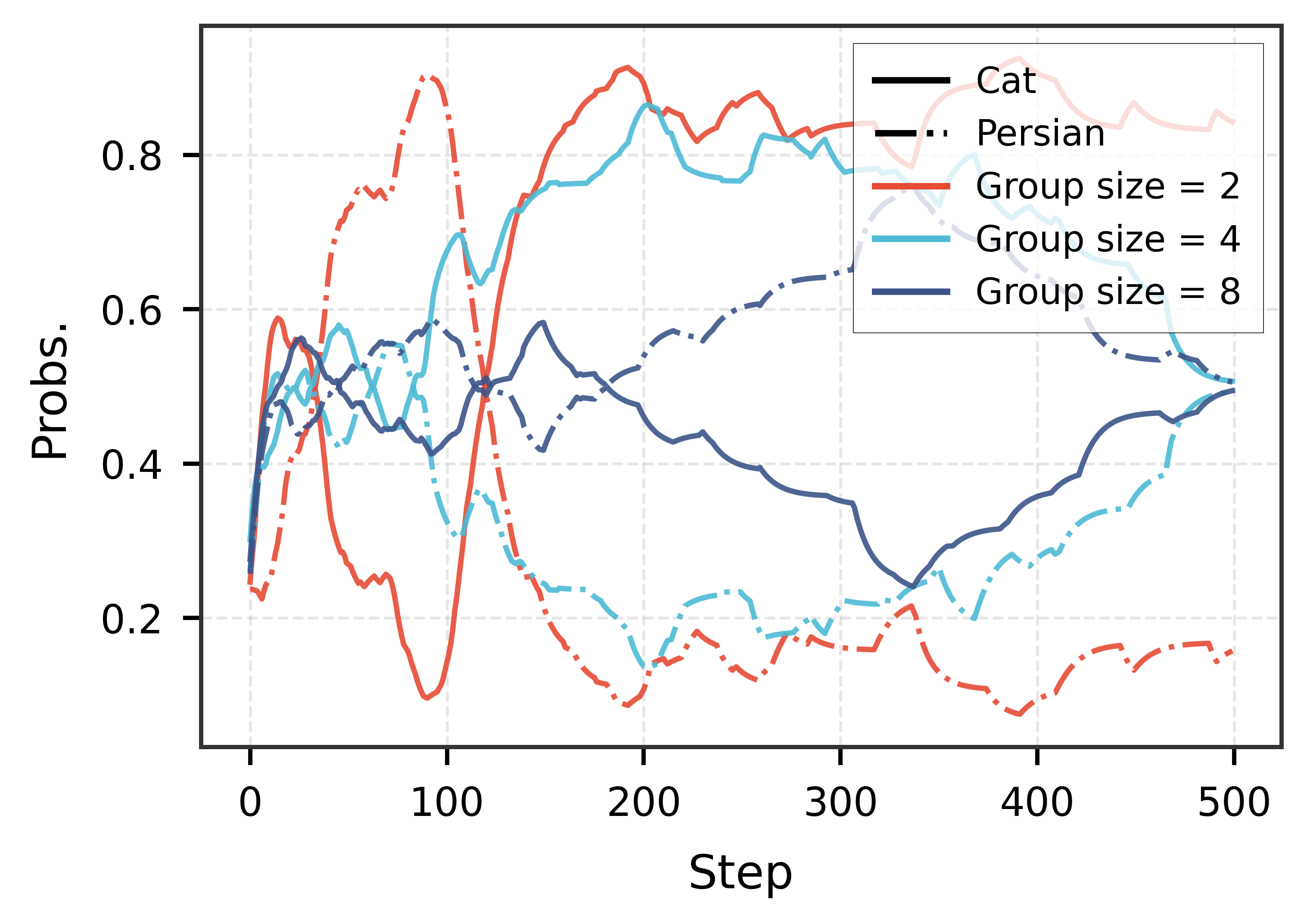}
        \caption{Inverse-success Calibration + Dist. Calib. in AdamW}
    \end{subfigure}
    \caption{We apply our proposed mitigations to the Softmax example. The left column (a, c) employs only inverse-Success advantage calibration, showing a delayed rate of collapse. The right column (b, d) incorporates distribution-level calibration (memory network, $\mu=0.5$), which arrests collapse and maintains diversity, even under the aggressive updates of AdamW.}
    \label{fig:toy_mitigation}
\end{figure*}

We validate the efficacy of our proposed calibration mechanisms using the toy setting described in \cref{sec2_4}.
\cref{fig:toy_mitigation} presents the evolution of probability mass.
We see that inverse-success advantage calibration alone significantly retards the rate of collapse compared to the uncalibrated baseline.
Coupling this with distribution-level calibration (memory network) enables the policy to reach a stable equilibrium where diversity is maintained, effectively arresting over-sharpening.
Moreover, despite AdamW's tendency to accelerate collapse, our methods still can prevent the aggressive elimination of alternative modes.

\section{Supplementary Experiment for \cref{sec4}}
\label{appendix_sup_exp_sec4}

\begin{table}[!th]
\caption{Comparison of AVG@8 performance in DeepSeek-R1-Distill-Qwen-1.5B trained on DAPO-Math-17K.}
\label{tab_supp_ds_model}
\centering
\begin{tabular}{@{}ccc@{}}
\toprule
Method & Math500 & O-bench \\ \midrule
GRPO   & 80.07   & 42.65   \\
DAPO   & 82.25   & 43.64   \\
IAC (Ours)   & 84.61   & 45.54   \\ \bottomrule
\end{tabular}
\end{table}

\begin{table}[!th]
\caption{Sensitivity of IAC to the calibration strength $\alpha$. We report AVG@8 and PASS@8 on Math500 and O-Bench for Qwen3-8B when varying $\alpha$ in IAC. When $\alpha=0$, IAC reduces to the vanilla GRPO objective. We also include a simple filtering baseline that discards the easiest and hardest 25\% of training queries (by empirical success rate), retaining only the middle 50\% of the difficulty spectrum.}
\label{tab:alpha_sensitivity}
\centering
\begin{tabular}{@{}cccccccc@{}}
\toprule
Metric                 & Dataset   & 0     & 0.5   & 1              & 1.5            & 2     & Filtering-based~\cite{diffcult_filter} \\ \midrule
\multirow{2}{*}{AVG@8}  & Math500 & 85.81 & 87.45 & \textbf{87.67} & 86.64          & 86.29 & 86.13           \\
                       & O-Bench & 49.79 & 50.47 & 50.49          & \textbf{52.11} & 50.17 & 50.21           \\ \midrule
\multirow{2}{*}{PASS@8} & Math500 & 91.39 & 93.79 & \textbf{94.54} & 94.37          & 93.85 & 92.06           \\
                       & O-Bench & 56.88 & 57.56 & 58.48          & \textbf{59.66} & 58.88 & 57.30           \\ \bottomrule
\end{tabular}
\end{table}

\textbf{Performance comparison across different datasets and models.}
We further evaluate IAC on DAPO-Math-17K~\cite{DAPO} using DeepSeek-R1-Distill-Qwen-1.5B to verify its generality.
As shown in \cref{tab_supp_ds_model}, IAC achieves the best performance, reaching 84.61\% on Math500 and 45.54\% on O-bench.
This represents consistent improvements over GRPO (80.07\% / 42.65\%) and DAPO (82.25\% / 43.64\%), demonstrating that IAC can effectively enhance mathematical reasoning capabilities when applied to a different backbone model and training corpus.

\textbf{Sensitivity to the calibration strength $\alpha$.}
\cref{tab:alpha_sensitivity} investigates how $\alpha$ in IAC affects downstream performance on Math500 and O-Bench.
Recall that $\alpha$ controls how strongly we up-weight advantages from hard, low-success queries and down-weight those from easy, high-success ones.
When $\alpha=0$, our method reduces exactly to GRPO, and the results at this setting coincide with the GRPO baseline.
As $\alpha$ increases from $0$ to $1$, we observe a consistent improvement on both datasets: AVG@8 and PASS@8 on Math500 peak at $\alpha=1$, while O-Bench attains its best PASS@8 at a slightly larger $\alpha=1.5$.
Beyond this moderate regime, further increasing $\alpha$ leads to mild degradation.
Intuitively, small $\alpha$ under-corrects sampling bias, allowing easy queries to keep dominating the update and driving over-sharpening.
In contrast, large $\alpha$ over-corrects, overly suppressing signal from already-mastered queries and making the optimization less sample-efficient.
As we can see, $\alpha=1$ appears to be a good choice in practice.

\cref{tab:alpha_sensitivity} also compares IAC against a filtering strategy that discards the easiest 25\% of training queries according to their empirical success rate~\cite{diffcult_filter}.
Across both Math500 and O-Bench, filtering yields only modest gains over GRPO and is consistently weaker than IAC.
This gap highlights a key limitation of hard filtering: removing easy queries entirely reduces the effective training distribution and discards useful gradient information, particularly about how to maintain correctness on already-solved patterns.
IAC, in contrast, keeps all queries in the optimization loop but rescales their influence on the update.
Easy queries contribute a smaller—but non-zero—signal, preventing them from overwhelming the batch partition function $Z'(q)$, while hard queries are emphasized in a smooth, data-dependent manner.
From a distributional perspective, filtering crudely prunes regions of the solution space, whereas IAC gently rebalances probability mass across difficulty levels.
The latter is more effective at preserving useful behaviors on easy instances while still allocating sufficient capacity to explore and stabilize difficult, low-success regimes, which explains the consistent empirical advantage of IAC over filtering-based baselines.

\begin{figure*}[!t]
    \centering
    \begin{subfigure}[b]{0.45\textwidth}
        \centering
        \includegraphics[width=\textwidth]{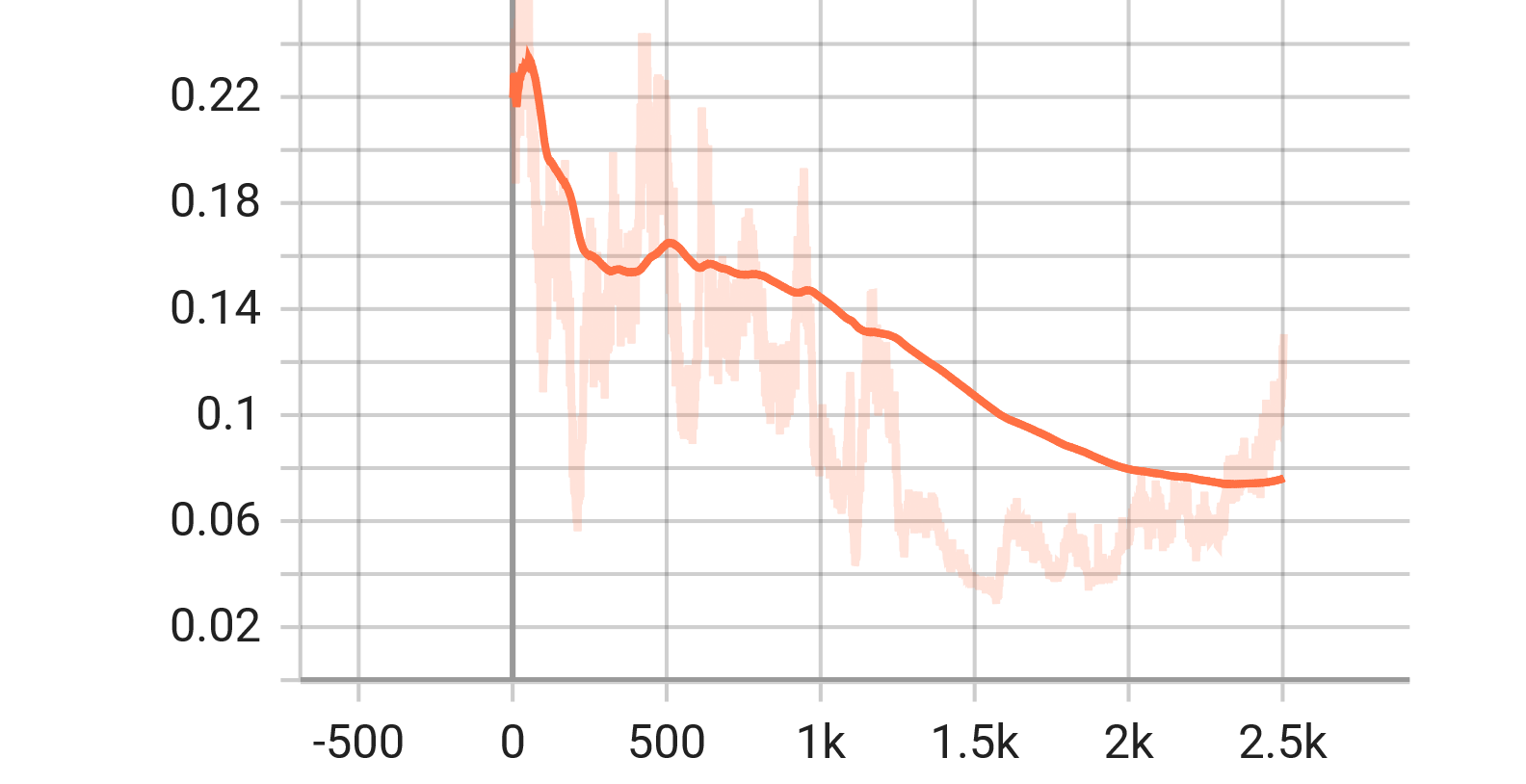}
        \caption{DAPO}
    \end{subfigure}
    \hfill
    \begin{subfigure}[b]{0.45\textwidth}
        \centering
        \includegraphics[width=\textwidth]{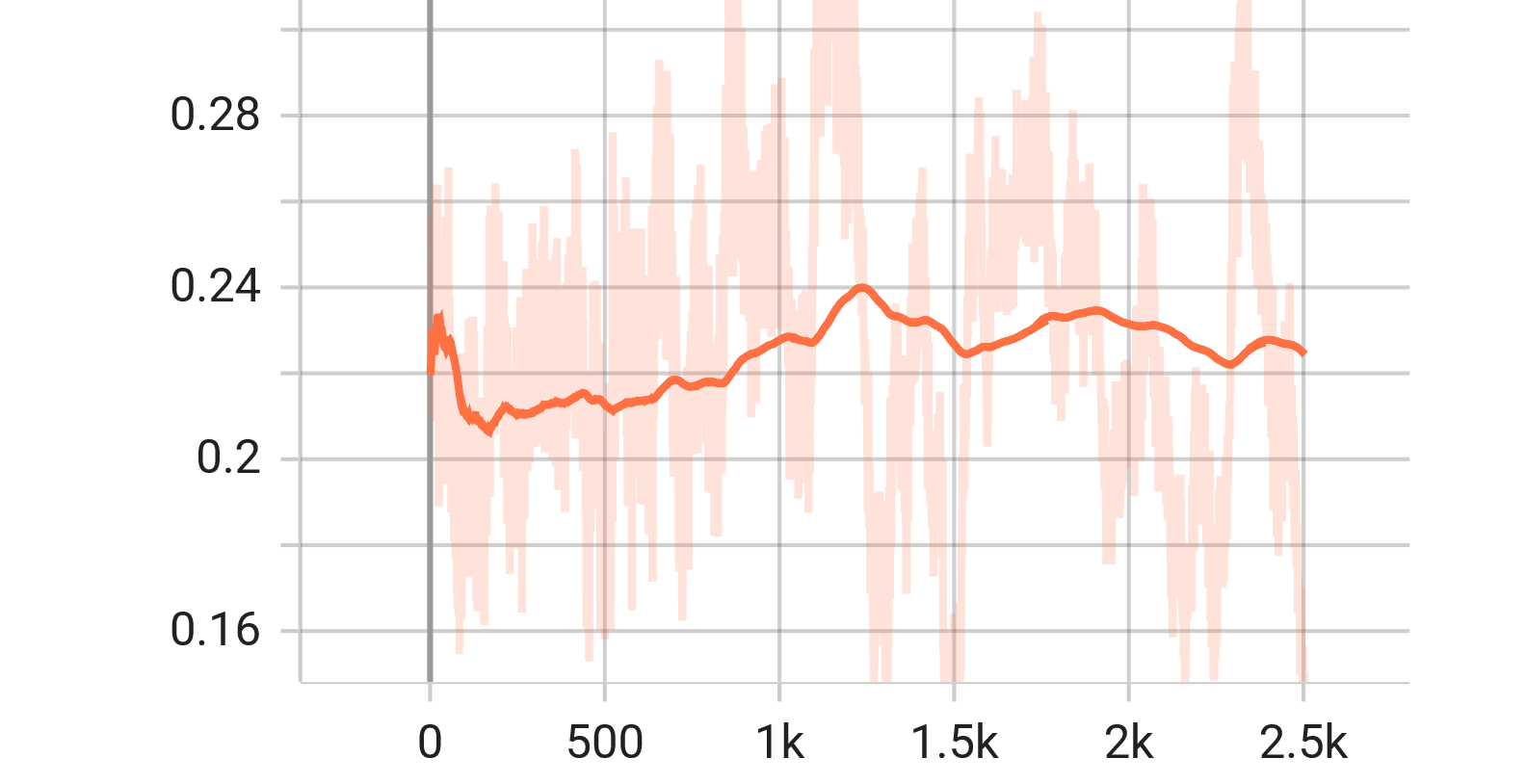}
        \caption{DAPO+IAC}
    \end{subfigure}
    \caption{Evolution of policy entropy under DAPO and DAPO+IAC.}
    \label{fig:entropy}
\end{figure*}

\textbf{Entropy evolution.}
\cref{fig:entropy} tracks the evolution of the policy entropy when applying DAPO alone versus DAPO combined with IAC.
Both methods start from similar initial entropy, but their trajectories diverge markedly over training.
Under plain DAPO, entropy decreases monotonically and eventually settles around $0.08$, signalling that the policy becomes almost deterministic.
In contrast, for DAPO+IAC, it stabilizes at a much higher level, fluctuating between $0.20$ and $0.24$ throughout training.
This indicates that the policy retains a substantial degree of stochasticity and continues to explore multiple semantic modes even late in optimization.


\end{document}